%% file: main.tex
\documentclass[pdflatex,sn-nature,iicol]{sn-jnl}
\usepackage{graphicx}
\usepackage{amsmath,amssymb}
\usepackage[title]{appendix}
\usepackage{xcolor}
\usepackage{booktabs}
\usepackage{adjustbox}
\usepackage{placeins}
\usepackage{tabularx}
\usepackage{array}
\usepackage{colortbl}
\usepackage{enumitem}
\usepackage{tikz}
\usepackage{pgfplots}\pgfplotsset{compat=1.18}
\usetikzlibrary{positioning,arrows.meta,calc}
\usepgfplotslibrary{groupplots}
\usepackage{xurl}
\hypersetup{pdftitle={Making every stage of medical device document assessment with large language models deterministic},pdfauthor={Ryosuke Suzuki},pdfsubject={Reproducibility of LLM-prepared assessments of medical device documents},pdfkeywords={large language models, reproducibility, determinism, medical devices, regulatory science}}
\newcommand{\ok}{\ensuremath{\checkmark}}

\begin{document}

\title[Deterministic assessment at every stage]{Making every stage of medical device document assessment with large language models deterministic}

\author*[1]{\fnm{Ryosuke} \sur{Suzuki}}\email{ryosuzuki@gflops-ai.com}

\affil*[1]{\orgname{GFLOPS Co., Ltd.}, \orgaddress{\city{Tokyo}, \country{Japan}}}

\abstract{To show that a medical device complies with regulation, experts judge whether documents and records meet each requirement and record the judgment for others to verify. For each assessment item that experts define, large language models could prepare a preliminary assessment: a label such as Satisfied or Not satisfied, a reason and quoted evidence. We sent fourteen commercial deployments of such models (seven models from three providers, through six routes) the same request repeatedly, with the deterministic sampling settings each accepted; every deployment returned differing assessments for at least one request, and in twelve the label changed. Reducing reasoning where the interface allows it did not reduce the variation, and prompts of retrieved passages removed it only for the smallest models. The same open-weight model was reproducible or not depending only on how it was served. A pipeline we built to be deterministic at every stage---text extraction and optical character recognition (OCR), chunking, embedding, indexing, retrieval and inference---returned byte-identical output at each stage across reruns, interruptions, edits, machines of one graphics processing unit (GPU) model and days, for blood glucose meter and infusion pump documents. OCR, embedding and approximate retrieval varied when their own serving was not made invariant, and outputs changed with the GPU model and with the inference engine's scheduling settings, which therefore belong in the recorded configuration. End-to-end determinism is achievable and gives a testable basis for audit and change control.}


\maketitle

\section*{Introduction}

Demonstrating that a medical device complies with regulation is expert work that must leave a record. Experts interpret the applicable requirements, judge whether the documents and records---technical files, test reports and data held in electronic quality management systems---meet them, and record each judgment with its evidence. The work recurs with every submission, design change and audit, on the manufacturer's side and on the side of notified bodies and regulators, and the records are kept for at least ten years \citep{eu2017mdr}; a notified body must document its conformity assessment so that its conclusions ``can represent objective evidence of such compliance to persons that are not themselves involved in the assessment'' \citep{eu2017mdr}, and the US Food and Drug Administration (FDA) reviews the substance of a 510(k) (premarket notification) submission only after a completeness screening \citep{fda2023estar,fdarta2022}.

Large language models (LLMs) could help with this work \citep{li2026scaling,fda2025elsa,ema2024llm}. We take a limited view of their role. Deciding which regulations, standards and guidance apply, and stating what must be shown as assessment items, remains the experts' work. For each item that experts define, an LLM can prepare a \emph{preliminary assessment}---a label such as Satisfied or Not satisfied, a short reason and the passages it relies on, quoted verbatim---for a qualified expert to check against the source; the judgment, and the responsibility for it, stay with the expert, whose decision, not the preliminary assessment, enters the quality record \citep{gangopadhyay2026safety}. Regulators' guidance recommends exactly this form of evidence, quoted sentences that can be checked \citep{ema2024llm}, and warns against the documented tendency of even expert users to over-rely on an automated suggestion \citep{parasuraman2010complacency,goddard2012automation,dratsch2023automation}.

The recorded judgment rests on what the expert was shown. A system that prepares such assessments should therefore return exactly the same assessment whenever the documents, the assessment items and its own configuration are the same. Software used in a quality management system must be validated for its intended use, including when it changes (ISO 13485:2016, subclause 4.1.6, incorporated by 21 CFR 820.10(a); \citealp{iso13485,cfr820,qmsr,fda2026csa}), and electronic-record systems for ``accuracy, reliability, consistent intended performance, and the ability to discern invalid or altered records'' \citep[\S 11.10(a)]{cfr11}; with a system whose output varies from run to run, a single difference observed after a change---to a document, an item, a model or the hardware---cannot be told from ordinary variation without repeated runs, and the individual assessments the change affects cannot be identified. An LLM pipeline does not give the same output for the same input by construction: floating-point reductions depend on how requests are batched \citep{he2025nondeterminism}, embeddings and retrieval drift with precision and with insertions \citep{wang2025reprorag}, a commercial deployment runs on hardware and software the user cannot see \citep{ong2026innovating}, and in a recent preprint covering five API providers, repeated outputs at temperature zero were character-identical 22\% of the time on average for the two models evaluated on every task \citep{rover2026sameprompt}. Reproducibility is a separate axis from accuracy, and this paper concerns only reproducibility.

Deterministic inference is now available: batch-invariant kernels and, more recently, verified and scheduling-based engines make a model's output independent of how requests are batched \citep{he2025nondeterminism,vllm_batch_invariance,qin2026vosti,zhao2026corun,gond2026llm42}. What has not been tested is whether a complete assessment pipeline, in which inference is one stage among several, can be made reproducible, and what a user of a commercial deployment actually observes in the fields of an assessment record. We asked whether preliminary assessments of public FDA documents can be produced again, byte for byte, and found three things. Commercial LLM deployments did not reproduce their assessments at the deterministic sampling settings they accepted: the quoted evidence changed more often than the label, one model varied differently through different routes, another changed labels only between days, and reducing reasoning did not help. For the open-weight models we served, reproducibility was a property of how the model was served. A pipeline deterministic from text extraction to inference reproduced every intermediate result and output across repetition, reordering, concurrency, interruption, corpus edits, machines and days for two device types, provided the GPU model and the inference engine's configuration are held fixed and recorded; left to their defaults, OCR, embedding and approximate retrieval were themselves sources of variation. We argue that, where judgments recorded in a quality system rest on LLM-prepared preliminary assessments, byte-level reproducibility of the entire pipeline is a strong and testable engineering condition for auditing those judgments and controlling change. It is not a regulatory requirement, and it is not sufficient for validation.

\section*{Results}

Each preliminary assessment has four outputs (Table~\ref{tab:example}): a label (here Satisfied, Not satisfied or Cannot determine), a comment giving the reason, the passages quoted as evidence, and the files they come from. We compared repeated runs of the same prompt on six fields in two layers. The \emph{substance} layer is what an expert relies on: the label, the cited files and the cited passages, where two runs cite the same passage when their quotations overlap at the same place in the same document. The \emph{text} layer is what the assessment record stores: the exact quotation strings, the comment and the full output, compared character for character. Table~\ref{tab:example} compares two of the five same-day runs of one prompt field by field; it illustrates how runs are compared, not which label is correct, since the study measures reproducibility, not accuracy.

\begin{table*}[t]
\centering
\caption{One prompt of Experiment~1: a preliminary assessment and how two of its runs are compared. Two of five runs of one prompt on the same day (\texttt{claude-opus-5-5}, day 1, runs 3 and 4; device K153561; texts shortened). The assessment item: ``At least two meters should be included in this study and at least 10 measurements should be taken per glucose level, per meter.'' Both runs cite the same document but different studies in it, and reach opposite labels; the other three runs answered \emph{Cannot determine}.}
\label{tab:example}
\small
\begin{tabularx}{\linewidth}{llXXXc}
\toprule
Layer & Field & What is compared & Run 3 & Run 4 & Same? \\
\midrule
Substance & Label & The model's assessment label (not a ground-truth label). & Satisfied & Not satisfied & no \\
& Files & The set of documents cited as evidence. & document 2 & document 2 & yes \\
& Passages & Where in the source the quotations are: two runs cite the same passage when their quotations overlap at the same place in the same document, whatever the spacing or order. & the repeatability and intermediate-precision studies: ``\ldots\ using three test strip lots and ten AG-607 blood glucose meters'' & the hematocrit and interference studies: ``The evaluation included 5 meters, each tested with two replicates \ldots'' & no \\
\midrule
Text & Quotes & The exact quotation strings, character for character. & 2 quotations & 3 quotations & no \\
& Comment & The exact text of the reason. & ``The precision studies used ten AG-607 meters \ldots\ This meets the minimum \ldots'' & ``Where the meter count and replicate numbers are reported, fewer than 10 measurements were taken \ldots'' & no \\
& Output & The whole response, byte for byte (not a separate output field). & & & no \\
\bottomrule
\end{tabularx}
\end{table*}

\subsection*{Commercial LLM deployments did not reproduce their assessments}
\label{sec:e1}

\begin{figure*}[t]
\centering
\includegraphics[width=\textwidth]{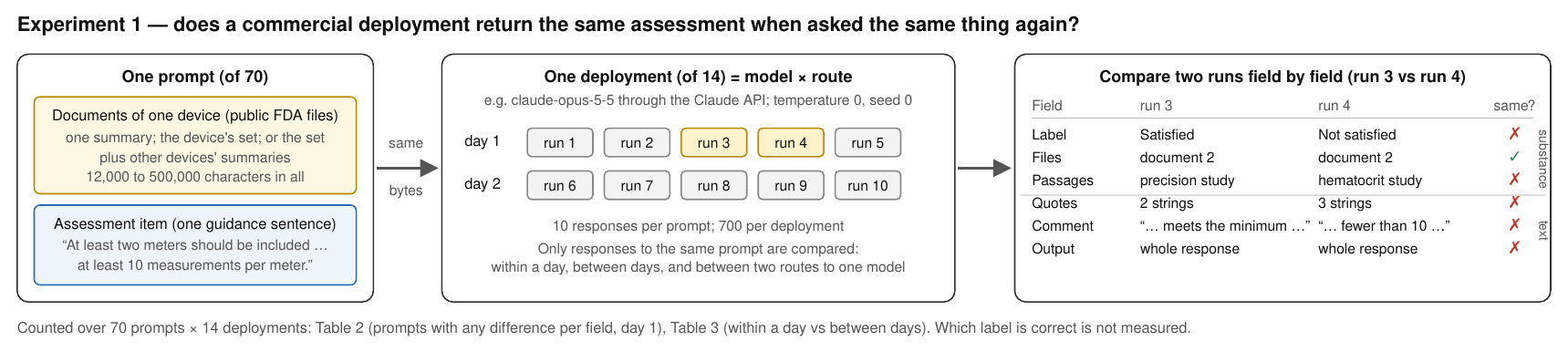}
\caption{Experiment~1. Each of 70 prompts, built from the public FDA documents of one device and one assessment item, is sent as a byte-identical request to each of 14 commercial deployments (a model reached through a route), five times on each of two days. Only responses to the same prompt are compared, on six fields in two layers; one pair of runs is shown (Table~\ref{tab:example} gives it in full). The experiment measures whether the same request gives the same assessment, not whether the assessment is correct.}
\label{fig:design1}
\end{figure*}

No commercial deployment returned the same output for every prompt, and twelve of fourteen changed at least one label (Table~\ref{tab:e1}; Supplementary Table~\ref{tab:e1days}). A \emph{deployment} is a model reached through a particular route: two OpenAI models through the OpenAI API and Azure OpenAI, three Claude models through the Claude API and Amazon Bedrock, and two Gemini models through the Gemini API and Google's Agent Platform (Experiment~1; Figure~\ref{fig:design1}). Each received 70 prompts built from public FDA documents for blood glucose meters, in three tiers of length (12 to 500 thousand characters), five times on each of two consecutive days, as byte-identical requests asking for temperature~0, seed~0 and schema-constrained output wherever accepted and reasoning left at the provider's default (Methods; Supplementary Table~\ref{tab:e1tiers} gives every metric by tier).

Within a single day, ten of the fourteen deployments changed the label of at least one prompt. The number of such prompts ranged from 0 to 18 of 70 by deployment, the largest being \texttt{gpt-6-luna} through Azure OpenAI (Table~\ref{tab:e1}). Two further deployments changed labels only between the two days, bringing the total to twelve (Supplementary Table~\ref{tab:e1days}). The sampling setting does not account for this. Eight deployments rejected temperature~0 and ran at their default sampling, the four Google deployments accept the field but, by Google's documentation, ignore it \citep{google_gemini36}, and the one model that ran at temperature~0 without reasoning tokens through both routes, \texttt{claude-haiku-4-5}, still changed its label on 4 of 70 prompts through the Claude API. Nor does the provider's default reasoning: the eight deployments whose interface allows reasoning to be reduced were re-measured at the lowest documented setting, which left reasoning tokens on almost no call, and the number of prompts on which the label changed rose in seven of the eight (for \texttt{gpt-6-luna}, from 14 to 42 of 70 through the OpenAI API and from 18 to 43 through Azure), while \texttt{gpt-6-luna} also reached the output limit without a parsable answer on 53 of its 1{,}400 calls (Supplementary Table~\ref{tab:e4reason}). Nor does the length of the prompt: when the fourteen deployments received the much shorter prompts that the pipeline described below sends to its own judgment model (six retrieved chunks, 7{,}200--8{,}700 characters, 49 prompts), nine of them still changed their full output on 48--49 of the 49 prompts and two on 1 and 5, and only three---the two Gemini models through the Gemini API and \texttt{claude-haiku-4-5} through Bedrock---returned every prompt byte-identically within a day (Supplementary Table~\ref{tab:e3prompts}).

\begin{table*}[t]
\centering
\caption{Experiment 1, within one day and with the same number of runs (each deployment's first measurement day, 5 runs per prompt, 70 prompts; dates in the footnote). Columns 3--8: prompts on which the field differed in at least one run; columns 3--5 are the substance layer and 6--8 the text layer (Table~\ref{tab:example}). Last column: probability that two runs of the same prompt differ in label (mean over prompts, 95\% bootstrap interval). ``Ours'' is the judgment stage of our deterministic pipeline (Gemma~4 31B on one NVIDIA RTX PRO 6000) receiving the same prompts. Rows are grouped by provider, with the two routes of a model adjacent; within a provider, models are in increasing order of the last column. Tables~\ref{tab:e1days}, S1 and S3 and Figure~\ref{fig:fields} use the same order. For rows in which no prompt changed its label the bootstrap interval is degenerate and is omitted (---); the exact one-sided 95\% upper limit on the share of prompts whose label changes is then 4.2\% (0 of 70).}
\label{tab:e1}
\small
\begin{adjustbox}{max width=\linewidth}
\begin{tabular}{llrrrrrrl}
\toprule
Route & Model & Label & Files & Passages & Quotes & Comment & Output & $P$(label differs) \\
\midrule
Ours (self-hosted) & gemma-4-31b & 0 & 0 & 0 & 0 & 0 & 0 & 0.000 --- \\
\midrule
Gemini API & gemini-3.8-flash & 0 & 0 & 0 & 0 & 14 & 14 & 0.000 --- \\
Agent Platform & gemini-3.8-flash & 0 & 0 & 0 & 0 & 11 & 11 & 0.000 --- \\
Gemini API$^\S$ & gemini-3.6-flash & 0 & 1 & 3 & 7 & 52 & 52 & 0.000 --- \\
Agent Platform$^\S$ & gemini-3.6-flash & 1 & 1 & 3 & 6 & 61 & 61 & 0.009 [0.000, 0.026] \\
\midrule
Claude API & claude-haiku-4-5 & 4 & 5 & 11 & 13 & 30 & 31 & 0.034 [0.009, 0.069] \\
Amazon Bedrock & claude-haiku-4-5 & 0 & 0 & 1 & 1 & 2 & 2 & 0.000 --- \\
Claude API & claude-opus-5-5 & 11 & 11 & 23 & 37 & 70 & 70 & 0.081 [0.039, 0.130] \\
Amazon Bedrock$^\ddagger$ & claude-opus-5-5 & 10 & 13 & 28 & 38 & 70 & 70 & 0.080 [0.034, 0.131] \\
Claude API$^\ddagger$ & claude-haiku-5-5 & 13 & 32 & 42 & 44 & 70 & 70 & 0.091 [0.049, 0.140] \\
Amazon Bedrock$^\ddagger$ & claude-haiku-5-5 & 14 & 36 & 42 & 46 & 70 & 70 & 0.104 [0.056, 0.156] \\
\midrule
OpenAI API & gpt-6-astra & 6 & 5 & 11 & 19 & 70 & 70 & 0.046 [0.014, 0.086] \\
Azure OpenAI$^\dagger$ & gpt-6-astra & 10 & 7 & 15 & 21 & 70 & 70 & 0.077 [0.034, 0.129] \\
OpenAI API & gpt-6-luna & 14 & 12 & 20 & 22 & 70 & 70 & 0.103 [0.057, 0.160] \\
Azure OpenAI$^\dagger$ & gpt-6-luna & 18 & 17 & 25 & 30 & 70 & 70 & 0.137 [0.086, 0.197] \\
\bottomrule
\end{tabular}
\end{adjustbox}

\smallskip
\footnotesize $^\dagger$ Measured on 7 October; $^\ddagger$ on 12 October; $^\S$ on 13 October; the other commercial routes on 4 October; Ours on 5 October.
\end{table*}

\begin{figure*}[t]
\centering
\input{fig2}
\caption{How often each field differed between runs of the same prompt (Experiment~1, day 1: 5 runs, 70 prompts): the number of prompts on which the field was not identical in all five runs, by deployment. Each row is one deployment; the three marks on a row are the label (circle), the cited passages (square) and the full output (triangle), joined by a line. Label and cited passages are the substance layer, full output the text layer. Counts and order are those of Table~\ref{tab:e1}, which also gives the probability that two runs differ, with intervals.}
\label{fig:fields}
\end{figure*}
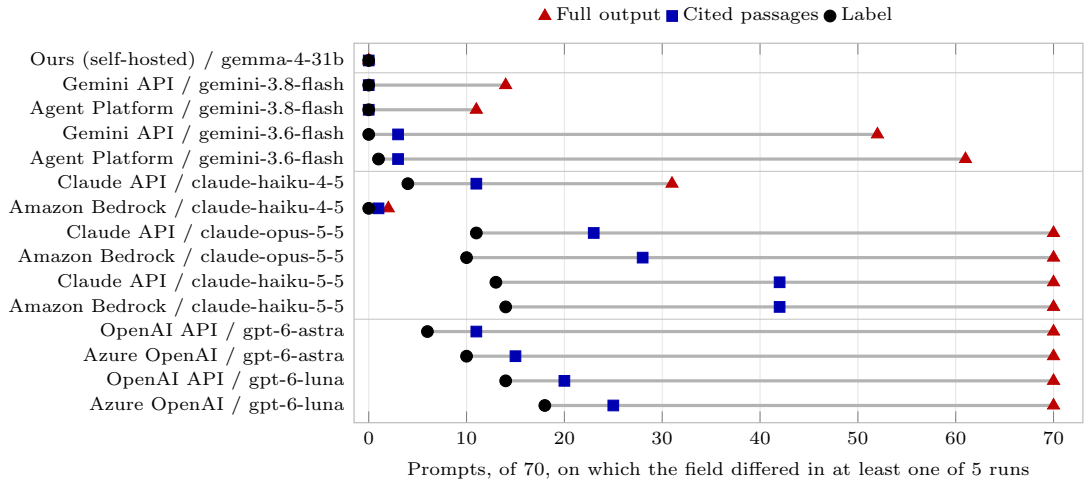

\paragraph{A stable label is not a stable record.} Quoted evidence varied more than labels, in substance and not only in text. In every deployment the number of prompts on which the cited passages differed was at least the number on which the label differed, and in twelve of the fourteen it was larger, by up to a factor of three (Table~\ref{tab:e1}; Figure~\ref{fig:fields}). \texttt{claude-opus-5-5} through the Claude API shows the pattern: its label changed on 11 of 70 prompts, but it cited a different passage---a different place in the documents, as in Table~\ref{tab:example}---on 23. Even on prompts whose label was the same in all five runs, the cited passages differed on between 0 and 30 prompts by deployment (Supplementary Table~\ref{tab:stable}). Formatting accounts for little of this: normalising white space, punctuation and case reduced the count of prompts with differing quotations by at most 6 in any deployment. The evidence, not the label alone, is what an expert reads and an auditor checks.

\paragraph{The same model could behave differently through different routes.} Each of the seven models was measured through two routes, and for one of them the routes differed. \texttt{claude-haiku-4-5}, with the same version and request settings, changed its label on 4 of 70 prompts, its cited passages on 11 and its full output on 31 through the Claude API, but on 0, 1 and 2 through Amazon Bedrock (exact McNemar test over the prompts: $p = 0.125$ for the label, $0.006$ for the passages and $< 10^{-6}$ for the output). For the other six models no route difference was detected at this resolution: the Claude 5.5 and OpenAI models varied in every output through both routes, and the two Gemini models differed between routes in output but not in label (Supplementary Table~\ref{tab:routes}). \texttt{claude-haiku-4-5} is also the one model that ran at temperature~0 without reasoning, so whether a route difference is visible depends on the sampling regime as well as on the model; either way, a validation record that names only the model does not identify what was validated.

\paragraph{Some label changes appeared only between days.} For thirteen of the fourteen deployments, label differences between the two days were no larger than within a day (Supplementary Table~\ref{tab:e1days}). The exception was \texttt{gemini-3.8-flash}: through both Google routes it gave the same label in all five runs of each day on every prompt, yet on one and two prompts the label of the second day differed from that of the first, under an unchanged model identifier, and the probability that two of its outputs differed rose from 0.01--0.09 within a day to 0.18--0.52 across days. A one-day validation of its labels would have called them stable. Prompt caching did not explain the variation either: through the OpenAI routes the probability that two runs differed was the same whether neither, one or both calls reported a cached prefix, and only through the Gemini API was it associated with cache state (Supplementary Table~\ref{tab:cache}).

\subsection*{Reproducibility depended on how the model was served}

The same open-weight model was reproducible or not depending only on how it was served. As a positive control, we sent the 70 prompts, on the same schedule, to the judgment stage of the deterministic pipeline described below (``Ours'' in Table~\ref{tab:e1}): Gemma~4 31B (FP8) served by vLLM with batch-invariant inference \citep{he2025nondeterminism,vllm_batch_invariance} on one NVIDIA RTX PRO 6000 GPU; all 700 outputs over two days were byte-identical, including prompts of about 130{,}000 tokens. A dedicated server sent one request at a time might be stable without any special measure, so we also sent the prompts from eight concurrent senders with batch invariance enabled and disabled, all else equal. Enabled, all 350 outputs were byte-identical to the sequential outputs; disabled, the full output varied on 58 of 70 prompts and the label on 5, and Qwen3-32B showed the same pattern, while Qwen3.8-27B could not be served in batch-invariant mode at all (Supplementary Table~\ref{tab:concurrency}). Batch invariance lengthened the concurrent run by 18\%. This changes one thing on one server, and is a direct form of the attribution that Rover et al.\ reached indirectly: variation comes from how the serving infrastructure executes a request, not from cloud hosting as such \citep{rover2026sameprompt}.

\subsection*{A complete assessment pipeline can be reproduced end to end}
\label{sec:e3}

\begin{figure*}[t]
\centering
\includegraphics[width=\textwidth]{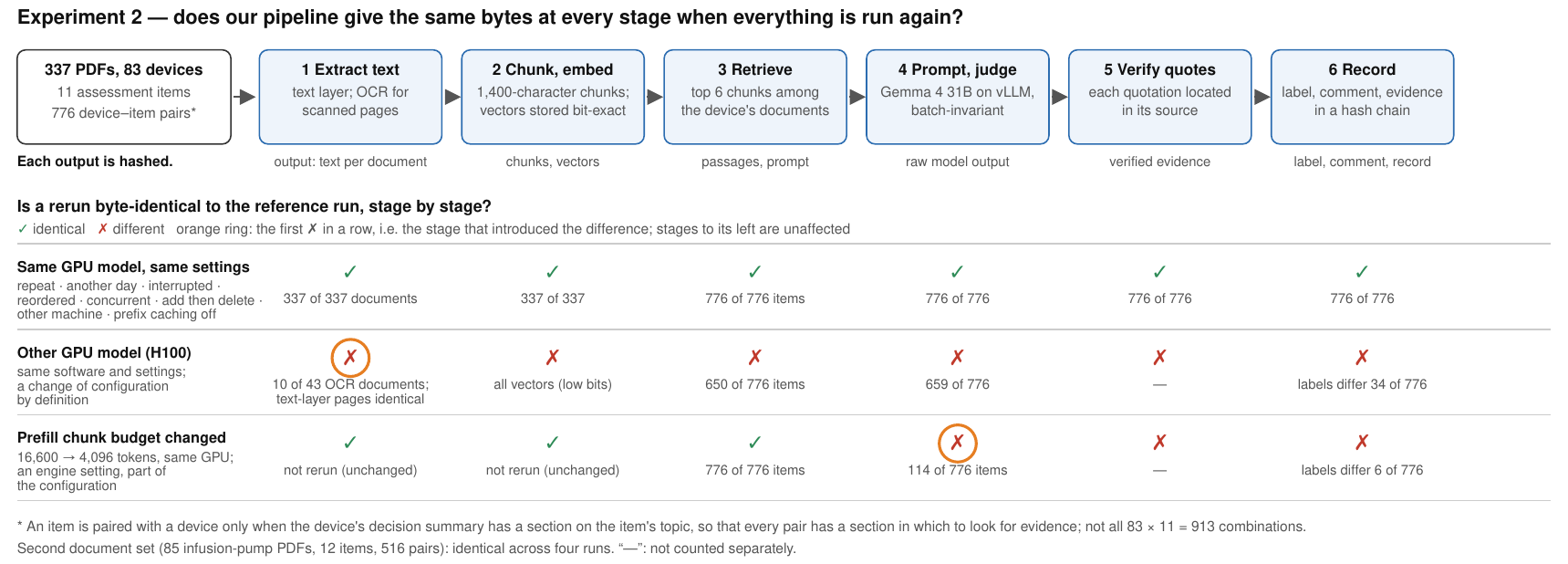}
\caption{Experiment~2. The pipeline extracts text, chunks and embeds it, retrieves the six best chunks among the device's documents, has the judgment model write the preliminary assessment, verifies each quotation against its source and records the result; the output of every stage is hashed. A reference run is compared, stage by stage, with reruns under operations that should not change the result and with runs under a changed configuration (another GPU model; another prefill chunk budget); the first stage that differs locates the cause. $^\ast$An item is paired with a device only when the device's decision summary has a section on the item's topic, so that every pair has a section in which to look for evidence (not all $83 \times 11 = 913$ combinations; Methods). The second document set (85 infusion-pump PDFs, 12 items, 516 pairs) was identical across four runs.}
\label{fig:design2}
\end{figure*}

Every intermediate result and output of a complete assessment pipeline was byte-identical across every operation we tried, for two medical device types, on one GPU model and one engine configuration (Table~\ref{tab:e3}); we write $\Theta$ for the configuration, everything other than the inputs that can change an output (Methods). The pipeline (Figure~\ref{fig:design2}) extracts text from each document, reading scanned pages by OCR, splits it into chunks, embeds each chunk, and stores chunks and vectors under hashes of their content; for each device--item pair it retrieves the six chunks that best match the item among that device's documents, scoring every chunk in scope exactly, builds the prompt, and has the judgment model (Gemma~4 31B, as above) write the preliminary assessment, whose quotations are then checked against the source by rule. One \emph{run} does this for every pair from an empty database and records a hash at every stage (Experiment~2).

We ran the pipeline on the 337 public FDA PDFs for blood glucose meters (3{,}150 pages, 100 of them read by OCR) and judged 776 device--item pairs over 83 devices. A reference run did this once, and we then repeated the whole process under operations that should not change the result---from fresh databases; interrupted and resumed during ingestion, including during OCR, and during judgment, with and without a restart of the inference server; with documents in reverse and in hash order; with 4 or 16 documents ingested and 1, 8 or 16 items judged concurrently; with a document added, assessed and deleted; on another machine with the same GPU model; and on a later day. Every per-stage hash matched the reference run for all 337 documents and all 776 assessments, and deleting the added document restored the database snapshot byte for byte. Determinism is not insensitivity to input: 19 of the 20 questions assessed for more than one device received different labels for different devices, and 746 of 776 comments are distinct. On infusion pumps---85 PDFs and 12 assessment items chosen by a rule fixed before the documents were fetched, 516 pairs---four runs in different orders, concurrencies and days agreed on every per-stage hash.

\begin{table*}[t]
\centering
\caption{Experiment 2. Agreement with the reference run, by stage: \ok\ = all 337 documents (ingestion: file, provenance, version and text hashes; embedding: chunk and vector hashes) or all 776 items (judgment: retrieval, prompt, raw output, label, comment, evidence, downgrade) identical; otherwise the count that matched. Operations in the first group should not change the result; the second changes the configuration $\Theta$, so differences are expected; the third removes one of the six properties.}
\label{tab:e3}
\footnotesize
\setlength{\extrarowheight}{0pt}
\begin{tabular}{p{7.5cm}>{\centering\arraybackslash}p{1.6cm}>{\centering\arraybackslash}p{1.8cm}>{\centering\arraybackslash}p{4.2cm}}
\toprule
Operation & Ingestion & Embedding & Judgment \\
\midrule
\multicolumn{4}{l}{\emph{Fixed configuration: operations that should not change the result}} \\
\arrayrulecolor{gray!45}\specialrule{0.3pt}{1pt}{1pt}
T1 Repeat; repeat on another day & \ok & \ok & \ok \\ \specialrule{0.3pt}{1pt}{1pt}
T2 Interrupt and resume; inference server restarted$^{a}$ & \ok & \ok & \ok \\ \specialrule{0.3pt}{1pt}{1pt}
T3 Reverse order; hash order & \ok & \ok & \ok \\ \specialrule{0.3pt}{1pt}{1pt}
T4 Concurrency 1, 4, 16 documents; 1, 8, 16 items & \ok & \ok & \ok \\ \specialrule{0.3pt}{1pt}{1pt}
T5 Add a document, assess, delete it & \multicolumn{2}{c}{snapshot identical} & \ok \\ \specialrule{0.3pt}{1pt}{1pt}
T6 Same GPU model, another machine & \ok & \ok & \ok \\ \specialrule{0.3pt}{1pt}{1pt}
T7$'$ Second H100 machine vs.\ the first H100 run & \ok & \ok & \ok \\ \specialrule{0.3pt}{1pt}{1pt}
T8/T9 Budget 16{,}384; prefix caching disabled & --- & --- & \ok \\ \specialrule{0.3pt}{1pt}{1pt}
Infusion-pump corpus, four runs$^{b}$ & \ok & \ok & \ok \\
\arrayrulecolor{black}\midrule
\multicolumn{4}{l}{\emph{Changed configuration $\Theta$: differences expected}} \\
\arrayrulecolor{gray!45}\specialrule{0.3pt}{1pt}{1pt}
T7 H100 80GB instead of RTX PRO 6000$^{c}$ & 327/337 & 0/337 & labels 742; raw 117/776 \\ \specialrule{0.3pt}{1pt}{1pt}
T8 Budget 4{,}096, 8 concurrent items, two runs$^{d}$ & --- & --- & raw 662, 654; labels 770, 768/776 \\ \specialrule{0.3pt}{1pt}{1pt}
T8 Budget 4{,}096, 16 concurrent items & --- & --- & raw 609; labels 768/776 \\ \specialrule{0.3pt}{1pt}{1pt}
T8 Budget 4{,}096, 1 concurrent item & --- & --- & \ok \\
\arrayrulecolor{black}\midrule
\multicolumn{4}{l}{\emph{One property removed (ablation on the reference GPU)}} \\
\arrayrulecolor{gray!45}\specialrule{0.3pt}{1pt}{1pt}
OCR without batch invariance, 16 concurrent docs$^{e}$ & 323/336 & --- & --- \\ \specialrule{0.3pt}{1pt}{1pt}
Embedding without batch invariance (vLLM pooling)$^{f}$ & --- & 0/337 & --- \\ \specialrule{0.3pt}{1pt}{1pt}
\quad control: pipeline embedding server, two concurrent runs & --- & \ok & --- \\
\arrayrulecolor{black}\bottomrule
\end{tabular}

\smallskip
{\footnotesize\raggedright T8 and T9 rerun judgment on the reference database (budget 16{,}600 tokens, prefix caching enabled). $^{a}$ Ingestion killed after 131 files and, separately, during OCR; judgment killed; the inference server killed 90 s into judgment and restarted. $^{b}$ Two sorted-order runs, one hash-order run with 16 concurrent documents and items, one on the next day. $^{c}$ The 294 text-layer documents matched and the 10 differing documents all contain OCR pages; labels agreed for 620 of the 650 items with identical prompts; an A100 80GB could not run the configuration (Methods). $^{d}$ The two runs agreed with each other on 633 of 776 raw outputs. $^{e}$ Stopped by a two-hour limit after 336 documents; OCR documents 29/42, text-layer documents 294/294. $^{f}$ The two concurrent runs agreed with each other on 3 of 337 documents. The hash-order run with 16 concurrent documents is the rerun with the corrected extractor (Supplementary Note~2).\par}
\end{table*}

\paragraph{Each GPU stage needs its own invariance.} Removing batch invariance from one stage at a time, on the reference GPU, moved that stage's output and no other (Table~\ref{tab:e3}). With the OCR servers run without batch invariance and sixteen documents ingested concurrently, the extracted text changed for 13 of 42 OCR documents and for none of the 294 text-layer documents. With embedding done by the inference engine's pooling runner without batch invariance, the vectors differed between a run made alone and two runs made concurrently for 332--336 of 337 documents, whereas the pipeline's embedding server, which computes one text at a time with deterministic kernels, gave identical vectors in two concurrent runs. Approximate retrieval indexes built from the same vectors varied with how they were built: hnswlib at its default search depth was reproducible only single-threaded in a fixed order, Elasticsearch's default vector index changed 3 of 33 queries on an identical rebuild, and its BM25 ranking changed only after a document was added and deleted (because deleted documents are retained as soft deletes) or the shard count changed (Supplementary Table~\ref{tab:baselines}); the pipeline's exact, scoped retrieval matched in every case.

\paragraph{The hardware class and the engine's scheduling belong in the configuration.} One run on an H100 instead of the RTX PRO 6000, with the same software and settings, did not reproduce the reference run, and the per-stage hashes located every difference: all 294 text-layer documents matched, OCR output differed for 10 of 43 OCR documents, every embedding set differed, the retrieved passages and prompts differed for 126 of 776 items, and labels matched for 742 of 776 items (95.6\%)---even among the 650 items whose prompts were identical, 30 labels changed, because greedy decoding is sensitive to small numerical differences \citep{yuan2025numerical}, as the vendors of GPU libraries document \citep{nvidia_cublas_repro,nvidia_cudnn_repro,pytorch_reproducibility,vllm_reproducibility}. This is a difference between GPU models, not variation on the H100: a second run on another H100 machine, eleven days later, reproduced the first H100 run exactly, every document, vector and judgment. The engine's prefill chunk budget behaved the same way, with a condition: with the budget lowered from 16{,}600 to 4{,}096 tokens, judgment at one item at a time was byte-identical to the reference, but with eight concurrent items the raw output differed on 122 and 114 items in two runs and on 143 items between them, and with sixteen on 167, while at 16{,}384 tokens, and with prefix caching disabled, every item was byte-identical (Table~\ref{tab:e3}; Vosti reported the same sensitivity of this batch-invariant mode to prefill chunking \citep{qin2026vosti}). The budget by itself changes nothing; when several prompts share a budget too small for them, how the scheduler divides their prefills across engine steps depends on which requests are present, and the batch-invariant kernels do not make the result independent of that division. The list of what changed and at which stage is the change-impact analysis a validation team needs when hardware or engine settings change, and both the GPU model and the engine's scheduling settings are therefore part of $\Theta$.

\section*{Discussion}
\label{sec:discussion}

We propose that, where LLMs prepare assessments for regulated work such as that of the medical device industry, byte-level reproducibility of the whole pipeline be a testable engineering condition: the same documents, assessment items and configuration should yield the same assessment, at every stage. The fourteen commercial deployments we measured did not provide this at the deterministic sampling settings they accepted, with reasoning at its default or at its minimum, and where the label held, the evidence behind it often changed; the open-weight models we served were reproducible or not depending on how they were served; and a pipeline built to be deterministic at every stage reproduced every intermediate result and output for two device types, on one hardware class and engine configuration. We measured the commercial deployments from the outside, as any customer would, so we do not attribute their variation to causes, and our results do not rank the models. They do show that a deployment's reproducibility cannot be established once and then assumed: it differed between models, for one model between routes, and for another between days, so that a test confined to one session would have missed it.

End-to-end determinism makes three things possible that agreement between runs alone does not. A stored assessment can be regenerated as well as kept: producing it again from the stored documents, items and configuration and obtaining the same bytes shows that the record is what the recorded configuration still produces for these inputs, and lets an auditor see the passages and prompt behind a judgment exactly as the expert saw them. The effect of a change can be listed: when a document, an item, a prompt template, a model or the hardware is changed and the assessments are run again, every difference is caused by that change, and the per-stage hashes show which assessments changed and at which stage, as the H100 run did, so that the judgments that relied on them can be looked at again---although a listed difference is not thereby a meaningful one (a change of GPU model alone moved 30 of 650 labels on identical prompts), nor is an unchanged assessment thereby unaffected. And when documents change, only the assessments whose retrieved evidence changed need the model again, since an unchanged input to the model has an unchanged output; retrieval is cheap and inference is not (Supplementary Table~\ref{tab:e1cost}).

What determinism does not give is correctness: a system that reproduces a wrong assessment exactly has reproduced a wrong assessment, and we have not measured how often the assessments are right. Two alternatives are weaker. Storing every output with an audit trail preserves what was produced, and is what remains once a hardware class is retired, but cannot show that the validated system still produces it, nor which later changes would alter it. Statistical consistency suffices for population-level questions \citep{li2026scaling,shyr2026repeatability}, and disagreement between runs is itself used as a signal of uncertainty \citep{wang2023selfconsistency,kuhn2023semantic,farquhar2024semantic,joshi2026chaos}; a recorded judgment, however, is an individual decision made on an individual assessment, and a deterministic pipeline can still estimate uncertainty by varying the configuration deliberately, as the H100 and chunk-budget runs did. For manufacturers choosing such tools today, the results suggest testing reproducibility directly, across days and on the quoted evidence, and recording the route, the engine settings and the hardware class, because the same model name covered different behaviour; without determinism, a vendor's silent change can be detected only statistically \citep{chen2024chatgpt,gao2025equality,chauvin2026logprob}, and served behaviour has changed under unchanged model identifiers \citep{anthropic_model_ids,anthropic_postmortem2025,blackwell2024uncertainty}. For auditors, whether the preliminary assessment behind a recorded judgment can be produced again is a concrete question, to be scaled to risk as Computer Software Assurance recommends \citep{fda2026csa}.

Determinism holds within a hardware class, so records outlive the hardware that produced them only as stored outputs. Technical documentation must be kept for at least 10 years, and 15 for implantable devices \citep{eu2017mdr,eu2017ivdr}, longer than a GPU model may be available and far longer than a hosted model version, which providers retire on months' notice \citep{msft_foundry_lifecycle,openai_deprecations,anthropic_deprecations}. Container images, model weights and drivers should therefore be archived with the records, and reference outputs re-established on new hardware by listing the changed assessments. This favours models an organisation serves itself, which today means open-weight models whose architecture the serving software supports deterministically; batch invariance cost 18\% in wall time in our test, and newer engines narrow the gap in some regimes \citep{zhao2026corun,qin2026vosti}.

Regulations and regulators' guidance do not require byte-level reproducibility, but they repeatedly ask that a recorded result can be checked, and industry guidance frames validation by risk \citep{gamp5,ispegampai}. Part~11 requires audit trails and reliable, consistent performance of electronic-record systems, applies to records that predicate rules require, here the expert's recorded judgment, and has been enforced with discretion since 2003 \citep{cfr11,fda2003part11}; the EU device regulations require the technical documentation to locate the evidence of conformity \citep{eu2017mdr,eu2017ivdr}; the FDA--EMA principles for AI in drug development ask for ``detailed, traceable, and verifiable'' documentation \citep[Principle 6]{fdaema2026gaip}; EU GMP Annex 11 requires life-cycle validation \citep{annex11}; and the EU AI Act requires high-risk systems, from 2028 for devices under notified-body assessment, to log events and to perform ``consistently in those respects throughout their lifecycle'' \citep{eu2024aiact,eu2026omnibus}, which we cite by analogy. Only a consultation draft of EU GMP Annex 22 names identical outputs, limiting its scope to ``models with a deterministic output'' and excluding LLMs from critical use \citep{annex22draft}, and its drafting group has since discussed admitting generative models under a risk-based strategy \citep{ema2026annex22report,ema2026iwgplan}. FDA's 2019 discussion paper defines a ``locked'' algorithm as one that gives the same result for the same input \citep{fda2019aiml}, and FDA's and PMDA's 2026 papers on generative-AI devices list reproducibility across repeated runs among the properties to evaluate \citep{fda2026genai,pmda2026genai}.

Previous work has made individual stages deterministic or measured variation, but not tested a complete assessment pipeline. Batch-invariant and tree-based kernels, verified replay and fixed execution shapes make inference independent of batching \citep{he2025nondeterminism,vllm_batch_invariance,zhang2026tbik,gond2026llm42,zhao2026corun,qin2026vosti}; ReproRAG finds embeddings and FAISS retrieval reproducible run to run under fixed seeds but displaced by insertions \citep{wang2025reprorag}, and deterministic-first retrieval, retrieval-aware ingestion and a finance-oriented harness address other stages without extraction or OCR \citep{cogotti2026noesis,bochenek2026ragdeterm,allu2026drac,khatchadourian2025drift}. Variation of hosted LLMs at temperature~0 has been measured in research settings \citep{atil-etal-2025-non,ouyang2025nondeterminism,rover2026sameprompt}, the same model has been found to differ between routes \citep{blackwell2024uncertainty,gao2025equality,cai2025substitution}, and providers describe seeds as ``best effort'' and temperature~0 as not ``fully deterministic'' \citep{openai_seed,anthropic_glossary}. LLMs have been validated for extraction from FDA device documents without repeated runs \citep{li2026scaling} and proposed for source-linked safety analysis \citep{gangopadhyay2026safety}. To our knowledge, no prior work tests byte-level reproducibility of a complete assessment pipeline from text extraction to quoted evidence, relates it to medical-device quality work, or measures, field by field on an assessment record, how commercial deployments vary across repeated requests, routes and days.

The study has limitations. It concerns reproducibility only; the assessment items are guidance sentences chosen by rule, and public 510(k) documents for two device types stand in for manufacturers' records. The pipeline used one judgment model family on a single GPU, relies on the batch-invariant mode of one inference engine, and its six properties were sufficient under the operations tried, which is an empirical finding, not a proof; no one outside the author's company has yet reproduced it. The commercial measurements cover two days per deployment, one client region and pay-as-you-go tiers, with eight deployments sampled at provider defaults; the prompts were whole documents, and the 70 prompts come from 49 device--item pairs resampled as independent units; the positive control, the concurrency test, the between-day count and the two addenda were added during the study (Supplementary Note~2). Usefulness is a property of assessments in the hands of experts, and this study establishes reproducibility, not usefulness: how often the assessments are correct, whether the quoted evidence is what an expert would cite, and whether review helps or biases the expert \citep{goddard2012automation,jabbour2023impact} are questions we cannot answer alone, and we publish this preprint in part to find the regulatory and quality-management experts with whom to answer them. With a deterministic pipeline, what the experts evaluate is fixed, and any later difference can be attributed to a change.

\section*{Methods}

\paragraph{Definitions.} An \emph{assessment item} is a requirement to be assessed, defined by the assessment's administrator, who also defines the label set. A \emph{preliminary assessment} is, for one item and a set of documents, a label from the label set, a comment giving the reason, and quoted evidence---passages copied verbatim from the documents, each with its cited file; the model's internal reasoning is not part of it. The \emph{assessment record} is the stored preliminary assessment with the hashes that identify its inputs and configuration; the \emph{quality record} is the expert's recorded judgment, which is what quality-system requirements and Part~11 concern, and we assume that an assessment record is retained with the judgment that relied on it. A \emph{route} is the way a model is reached (a provider's API, a cloud platform or a self-hosted server) and a \emph{deployment} is a model reached through a particular route. The \emph{configuration} $\Theta$ is everything other than the inputs that can change an output: the extraction, chunking and embedding components, the retrieval parameters, the prompt template and label set, the model weights, the inference engine's version and settings, including its prefill chunk budget and concurrency limit, and the hardware class (GPU model with its driver and CUDA versions); the inputs are the document bytes and the assessment items, and a \emph{manifest} is the recorded listing of $\Theta$. \emph{Byte-level reproducibility} means that, for a fixed $\Theta$, every intermediate result and output is identical byte for byte across runs that differ only by operations that should not matter: repeating, restarting, interrupting and resuming, reordering, changing concurrency, and adding and deleting a document. We use \emph{reproducibility} in the computational sense; in the terms of FDA's draft guidance on AI-enabled device software \citep{fda2025aidsf}, repeating a request under the same conditions tests repeatability and agreement across days, orders, concurrency and machines tests reproducibility under pre-specified changed conditions.

\paragraph{Documents and assessment items.} The main corpus is the public FDA documents for over-the-counter blood glucose test systems (product code NBW; in vitro diagnostic devices for self-testing in the EU) with a decision date from 2015 to 2025 in openFDA and a published 510(k) summary: 133 510(k) summaries, 132 decision summaries and 72 predicate summaries, 337 PDFs. The assessment items are the 11 sentences of FDA's 2020 guidance for these systems that state a numeric minimum of a counted item (``at least'' or ``a minimum of'' followed by a numeral and a count of meters, strips, vials, lots, days, measurements or concentrations). An item is paired with a device when the device's decision summary has a section whose heading matches the item's topic, so that every pair has a section in which to look for evidence; 83 devices had such a section, giving 776 pairs, of which the reference run labelled 356 \emph{Cannot determine} (324 by the model and 32 by the verification rule), 380 \emph{Satisfied} and 40 \emph{Not satisfied}. We restricted the pairs in this way because the hard case for reproducibility is a judgment that has to find and quote evidence. The second corpus is the 43 FDA 510(k) summaries for infusion pumps (product code FRN) cleared in 2015--2025 and the 42 predicate summaries they cite, 85 PDFs; its 12 items are every 8.9th of the 107 sentences containing ``should'' or ``recommend'' in Section~5 of FDA's infusion pump guidance (Supplementary Table~\ref{tab:frn}), each assessed for every device (516 pairs). Both pairing rules were fixed before any assessment.

\paragraph{Experiment 1: commercial deployments.} The 70 prompts were built from the 776 pairs, ordered by 510(k) number and item, by a fixed rule without randomness: every 38th pair for the 20 \emph{short} prompts (the 510(k) summary; 12--47 thousand characters), every 19th for the 40 \emph{mid} prompts (the device's document set; 31--119 thousand) and every 77th for the 10 \emph{long} prompts (the set plus other devices' summaries; 483--502 thousand), skipping pairs whose summary had under 1{,}000 characters of text; the series overlap, so the 70 prompts come from 49 distinct pairs. Each prompt was sent five times on each of two consecutive calendar days (JST), one request at a time, from a virtual machine in Google Cloud us-central1; the days differ by deployment (Supplementary Table~\ref{tab:e1routes}), and the same-day route comparisons re-sent the OpenAI API's short prompts and \texttt{claude-opus-5-5} through the Claude API on the days of their counterparts. The size of the experiment was set by cost; five identical runs exclude only a per-run change probability above 45\% at the one-sided 95\% level, so a zero in Table~\ref{tab:e1} means that no change was observed, not that the deployment is stable. Requests were sent by Python's standard HTTP client without vendor SDKs, asking for temperature~0, seed~0 and JSON-schema output wherever accepted, with no reasoning parameter and an output limit of 16{,}384 tokens; settings a provider rejected were removed and recorded. Both OpenAI models, and \texttt{claude-opus-5-5} and \texttt{claude-haiku-5-5}, rejected temperature; the Gemini deployments accept the field but ignore it \citep{google_gemini36}. Reasoning tokens were reported on every call of the Gemini deployments and of \texttt{gpt-6-luna}, on most calls of \texttt{gpt-6-astra} and the Claude 5.5 models, and on none of \texttt{claude-haiku-4-5}. Of the calls reported, 78 to the Claude API failed when a credit balance ran out and were resent, 12 to Azure were retried after a rate-limit response, and 39 of \texttt{gpt-6-luna} and \texttt{gpt-6-astra} reached the output limit without parsable JSON and count as differing in every field. The system prompt, the user prompt format and the JSON schema are reproduced in Supplementary Note~1; the 70 prompts, their hashes and every response with its headers, timestamps and usage fields are in the supplementary material, and Supplementary Table~\ref{tab:costdates} gives every deployment, date, start and end time and estimated cost (\$1{,}291 in all, from provider-reported tokens at list prices).

\paragraph{Experiment 1 addenda.} \emph{Reduced reasoning}: on 14 and 15 October the 70 prompts were sent again to \texttt{gpt-6-luna} through both routes with \texttt{reasoning\_effort: "none"}, to \texttt{claude-haiku-5-5} through both routes with \texttt{thinking: \{type: "disabled"\}}, and to the four Gemini deployments with the lowest documented thinking level (\texttt{LOW} for \texttt{gemini-3.8-flash}, \texttt{MINIMAL} for \texttt{gemini-3.6-flash}); \texttt{gpt-6-astra} and \texttt{claude-opus-5-5} reject these settings. No setting was rejected. \emph{Retrieved-passage prompts}: for the 49 pairs, the chat messages that the pipeline's reference run sent to its judgment model were rebuilt from the chunk export and the recorded retrieval results, verified against the recorded prompt hash of every item, and sent five times on each of the same two days to all fourteen deployments at the settings above; two calls of \texttt{claude-opus-5-5} through the Claude API failed and were omitted.

\paragraph{Concurrency test.} The 70 prompts were split into eight groups sent concurrently, five runs each, to one vLLM 0.29.0 server on one RTX PRO 6000 with batch-invariant inference enabled or disabled and all else equal, for Gemma~4 31B (FP8), Qwen3-32B (BF16; 57 prompts fit its context) and Qwen3.8-27B (FP8), the Qwen models with thinking disabled and the attention kernel fixed (\texttt{TRITON\_ATTN}).

\paragraph{Experiment 2: pipeline.} Six properties fix what each stage does. \emph{Content addressing}: every document version, extraction, chunk set, embedding set, retrieval result, prompt and raw output is identified by the SHA-256 hash of its content and of the producing component's manifest, which lists that component's models, libraries and parameters, not the pipeline's source code. \emph{Deterministic extraction}: the same file yields the same text, including pages read by OCR with a self-hosted model under batch-invariant inference and greedy decoding. \emph{Deterministic embedding}: for a fixed configuration the same text yields the same chunks and the same vectors bit for bit whenever they are computed; within one database, vectors are stored under their content hash and reused. \emph{Exact, scoped retrieval}: the result depends only on the query and the documents in scope. \emph{Serving-invariant judgment}: for a fixed engine configuration, the model's output depends only on the prompt, not on which other requests are served with it, their order, or whether cached prefixes are reused; implemented with batch-invariant inference, greedy decoding and grammar-constrained output. \emph{Deterministic records}: quotations are verified against the source by rule, unverifiable ones removed and the label downgraded by rule, and every run is written to a hash-chained ledger. The implementation tested is AskDona Assessment, a product of the author's company (Competing interests). Pages whose text layer has fewer than 200 characters are rendered at 150 dpi and read by olmOCR-2, with GLM-OCR used when olmOCR-2's output is truncated; chunks of at most 1{,}400 characters are embedded one at a time with Qwen3-Embedding-8B; retrieval returns the top six of the device's chunks (4{,}977 in the main corpus) by BM25 and cosine similarity combined; judgment uses Gemma~4 31B (FP8) on vLLM 0.29.0 with batch invariance; and a quotation that is not found in its cited passage is removed, the label being downgraded to \emph{Cannot determine} only when no quotation remains (41 quotations and 32 labels in the reference run). Every component, version, parameter, rule and hash is listed in Supplementary Table~\ref{tab:params}. Judging the 776 pairs took 20.3 minutes with eight concurrent items.

\paragraph{Experiment 2: operations.} Each run starts from an empty database, re-reads every OCR page and re-embeds every chunk, and records a hash at every stage; judgment-only reruns reuse the reference database's chunks and vectors. The operations were: T1, a repeat from fresh databases, also on a later day; T2, ingestion killed after 131 files and, separately, after 3 files during OCR, then resumed, judgment killed and rerun, and the inference server killed 90 seconds into judgment and restarted before the rerun; T3, sorted, reverse and hash order; T4, 1, 4 and 16 concurrent documents and 1, 8 and 16 concurrent items; T5, a document added, assessed with the retrieval scope set to the whole corpus, and deleted; T6, the same GPU model on another machine; T7, an H100 80GB (an A100 could not run the configuration), run on 4 October and again on 14 October on a second H100 machine created from a disk image of the reference machine, with the GPU memory share raised to give the same absolute memory; T8, judgment rerun with the prefill chunk budget (\texttt{max-num-batched-tokens}) changed from 16{,}600 to 4{,}096 (twice with 8 concurrent items, once each with 1 and 16) and to 16{,}384 tokens; T9, judgment rerun with prefix caching disabled. Two ablations removed batch invariance from one stage: the OCR servers started with \texttt{VLLM\_BATCH\_INVARIANT=0} and the corpus ingested with 16 concurrent documents (stopped by a two-hour limit after 336 of 337 documents, which are the documents compared), and the corpus embedded by vLLM's pooling runner for the same model without batch invariance, once alone and twice concurrently, with the pipeline's embedding server run twice concurrently as the control.

\paragraph{Baselines.} From the same 4{,}977 chunks, vectors and 33 query embeddings we built hnswlib 0.8.0 indexes ($M = 16$, ef\_construction $= 200$, default search depth ef $= 10$) and Elasticsearch 9.5.4 indexes (default mapping, one shard) under the operations of Supplementary Table~\ref{tab:baselines}, plus a sweep of the hnswlib search depth over 10, 24, 50, 100, 200 and 500 (at ef $\geq 100$ every rebuild matched the first on all 33 queries, with recall 1.0 against exact search) and a soft-delete test in which setting \texttt{index.soft\_deletes.retention\_lease.period} to zero and force-merging restored all 33 BM25 scores.

\paragraph{Measures and statistics.} For each prompt we determine whether two runs differ in each of six fields: the label; the set of cited files; the cited passages, where each quotation is located in its source document after normalising white space and two runs agree if every quotation of each overlaps, by at least one character, a quotation of the other in the same document (a lenient, non-transitive rule, so the Passages counts are lower bounds); the exact quotation strings; the comment; and the output bytes. Files, passages and quotations are compared as sets; a call that returned no parsable JSON differs from every other run in every field; between 88\% and 100\% of quotations were located in the source, by deployment (Supplementary Table~\ref{tab:stable}); Experiment~1 compares raw model outputs without the pipeline's verification rule. A prompt counts as differing in a field when any pair of its runs differs; such counts cannot fall as runs are added. The probability that two runs differ is computed per prompt over all pairs of runs and averaged over prompts, with 95\% intervals from 2{,}000 percentile bootstrap resamples of prompts (seed 0). The normalised comparison removes white space, punctuation and case before comparing. Between days, a prompt counts as changed when its label was the same in all runs of each day but differed between the days. Route pairs are compared with exact McNemar tests on the paired prompt-level indicators. Experiment~2 compares, for every document and item, the per-stage hashes of each run with those of the reference run.

\paragraph{Software and data.} Experiment~2 ran on Ubuntu 22.04.5 with NVIDIA driver 580.178.04 (CUDA 13.0), vLLM 0.29.0 with \texttt{VLLM\_BATCH\_INVARIANT=1} (a beta feature in this version), \texttt{RedHatAI/gemma-4-31B-it-FP8-dynamic} (FP8 weights and KV cache, enforce-eager, chunked prefill, xgrammar structured output), \texttt{allenai/olmOCR-2-7B-1025}, \texttt{zai-org/GLM-OCR} and \texttt{Qwen/Qwen3-Embedding-8B} in PyTorch with deterministic algorithms; revisions and full environment records for every run, the document lists with hashes (\texttt{documents.csv}, \texttt{frn\_documents.csv}; all files were downloaded again from FDA and were identical), and the per-stage hashes are in the supplementary material. The measures, operations and conditions were fixed before the runs and every change was recorded when it was made; Supplementary Note~2 lists them all.
\backmatter

\bmhead{Data availability}
All inputs, prompts, outputs and per-stage hashes, with the document manifests (\texttt{documents.csv}, \texttt{frn\_documents.csv}) and the analysis scripts, will be deposited at Zenodo under the reserved DOI 10.5281/zenodo.23282335 on publication of this preprint; the documents themselves are public at FDA.

\bmhead{Code availability}
The implementation tested, AskDona Assessment (commit 7adab97 with the extractor change described in Methods), is proprietary; it will be made available to editors and reviewers on journal submission, and a minimal reference implementation of the six properties, with a script that checks the deposited per-stage hashes on an NVIDIA RTX PRO 6000, will be released with the journal version under an open licence. The analysis scripts that produce every number in this paper from the deposited records are in the supplementary material.

\bmhead{Acknowledgements}
This study was funded by GFLOPS Co., Ltd., which employs the author. Two colleagues at GFLOPS reviewed drafts of the manuscript and checked its facts. Large language model tools (Anthropic's Claude, through Claude Code) were used under the author's direction to write and check analysis scripts and to draft and edit text; the author verified every number and statement and is responsible for the content.

\bmhead{Competing interests}
The author is an employee of GFLOPS Co., Ltd., which develops and sells AskDona Assessment, the implementation evaluated here, and is the inventor on a provisional patent application covering the pipeline described.

\bmhead{Author contributions}
R.S. designed the study, built the pipeline, ran the experiments, analysed the data and wrote the manuscript.

\bibliography{refs}

\clearpage
\section*{Supplementary Information}
\setcounter{table}{0}\renewcommand{\thetable}{S\arabic{table}}\setcounter{figure}{0}\renewcommand{\thefigure}{S\arabic{figure}}
Supplementary Tables S1--S3 give the metrics over all runs of Experiment~1, the routes and accepted settings, and the tokens and estimated costs; Tables S4--S10 give the same-day route comparisons, the comparison by cache status, the label-stable prompts and located quotations, the concurrency test, the calls and costs by deployment and date, the infusion-pump items and the label transitions of the H100 and chunk-budget runs; Figure S1 gives the probabilities of Tables~2 and~3 with intervals; Tables S11 and S12 give the retrieved-passage and reduced-reasoning measurements of 14--15 October; Table S13 the across-day results of Experiment~1; Table S14 the retrieval baselines; Table S15 every component and parameter of the pipeline; Supplementary Note~1 the prompts and schema; and Supplementary Note~2 the changes made during the study. The supplementary files listed in Methods (prompt index and prompts, document manifests, per-stage hashes and analysis scripts) will be deposited at Zenodo (DOI 10.5281/zenodo.23282335).

\begin{sidewaystable*}
\centering\scriptsize
\caption{Experiment 1 by route, model and tier, computed over all runs of both days (10 runs per prompt). Azure OpenAI was measured on 7--8 October, Bedrock \texttt{claude-opus-5-5} and both \texttt{claude-haiku-5-5} deployments on 12--13 October, both \texttt{gemini-3.6-flash} deployments on 13--14 October, the other commercial runs on 4--5 October. $P$: probability that two runs of the same prompt differ (mean over prompts). Any: prompts with at least one difference. Latency in seconds (median).  Rows are in the order of Table~\ref{tab:e1}, with the tiers of each deployment together.}
\label{tab:e1tiers}
\begin{adjustbox}{max width=\linewidth}
\begin{tabular}{lllrrrrrrrrr}
\toprule
Route & Model & Tier & Prompts & Runs & $P$ label & $P$ files & $P$ quotes & $P$ output & Any label & Any output & s \\
\midrule
Ours & gemma-4-31b & short & 20 & 10 & 0.000 & 0.000 & 0.000 & 0.000 & 0 & 0 & 3.9 \\
Ours & gemma-4-31b & mid & 40 & 10 & 0.000 & 0.000 & 0.000 & 0.000 & 0 & 0 & 9.74 \\
Ours & gemma-4-31b & long & 10 & 10 & 0.000 & 0.000 & 0.000 & 0.000 & 0 & 0 & 79.85 \\
\midrule
Gemini API & gemini-3.8-flash & short & 20 & 10 & 0.000 & 0.000 & 0.000 & 0.048 & 0 & 3 & 2.64 \\
Gemini API & gemini-3.8-flash & mid & 40 & 10 & 0.014 & 0.014 & 0.014 & 0.105 & 1 & 9 & 3.69 \\
Gemini API & gemini-3.8-flash & long & 10 & 10 & 0.000 & 0.000 & 0.000 & 0.451 & 0 & 10 & 5.33 \\
\addlinespace
Agent Platform & gemini-3.8-flash & short & 20 & 10 & 0.000 & 0.000 & 0.000 & 0.360 & 0 & 14 & 4.24 \\
Agent Platform & gemini-3.8-flash & mid & 40 & 10 & 0.028 & 0.028 & 0.083 & 0.298 & 2 & 22 & 6.67 \\
Agent Platform & gemini-3.8-flash & long & 10 & 10 & 0.000 & 0.000 & 0.000 & 0.238 & 0 & 5 & 19.85 \\
\addlinespace
Gemini API & gemini-3.6-flash & short & 20 & 10 & 0.000 & 0.000 & 0.000 & 0.071 & 0 & 4 & 4.14 \\
Gemini API & gemini-3.6-flash & mid & 40 & 10 & 0.000 & 0.013 & 0.052 & 0.484 & 0 & 40 & 5.81 \\
Gemini API & gemini-3.6-flash & long & 10 & 10 & 0.000 & 0.000 & 0.093 & 0.913 & 0 & 10 & 9.82 \\
\addlinespace
Agent Platform & gemini-3.6-flash & short & 20 & 10 & 0.028 & 0.028 & 0.056 & 0.486 & 1 & 20 & 3.12 \\
Agent Platform & gemini-3.6-flash & mid & 40 & 10 & 0.009 & 0.009 & 0.066 & 0.472 & 1 & 40 & 5.20 \\
Agent Platform & gemini-3.6-flash & long & 10 & 10 & 0.000 & 0.000 & 0.000 & 0.569 & 0 & 10 & 6.91 \\
\midrule
Claude API & claude-haiku-4-5 & short & 20 & 10 & 0.046 & 0.046 & 0.112 & 0.279 & 2 & 11 & 2.2 \\
Claude API & claude-haiku-4-5 & mid & 40 & 10 & 0.013 & 0.026 & 0.131 & 0.356 & 1 & 21 & 3.67 \\
Claude API & claude-haiku-4-5 & long & 10 & 10 & 0.047 & 0.129 & 0.171 & 0.429 & 1 & 6 & 3.21 \\
\addlinespace
Bedrock & claude-haiku-4-5 & short & 20 & 10 & 0.000 & 0.000 & 0.000 & 0.023 & 0 & 1 & 2.52 \\
Bedrock & claude-haiku-4-5 & mid & 40 & 10 & 0.000 & 0.000 & 0.012 & 0.017 & 0 & 2 & 3.84 \\
Bedrock & claude-haiku-4-5 & long & 10 & 10 & 0.000 & 0.000 & 0.000 & 0.000 & 0 & 0 & 5.0 \\
\addlinespace
Claude API & claude-opus-5-5 & short & 20 & 10 & 0.061 & 0.033 & 0.138 & 1.000 & 3 & 20 & 3.8 \\
Claude API & claude-opus-5-5 & mid & 40 & 10 & 0.101 & 0.146 & 0.427 & 0.999 & 10 & 40 & 6.66 \\
Claude API & claude-opus-5-5 & long & 10 & 10 & 0.036 & 0.056 & 0.453 & 1.000 & 1 & 10 & 7.4 \\
\addlinespace
Bedrock & claude-opus-5-5 & short & 20 & 10 & 0.033 & 0.000 & 0.096 & 1.000 & 2 & 20 & 5.22 \\
Bedrock & claude-opus-5-5 & mid & 40 & 10 & 0.117 & 0.135 & 0.432 & 0.998 & 12 & 40 & 10.79 \\
Bedrock & claude-opus-5-5 & long & 10 & 10 & 0.000 & 0.000 & 0.318 & 1.000 & 0 & 10 & 13.68 \\
\addlinespace
Claude API & claude-haiku-5-5 & short & 20 & 10 & 0.107 & 0.153 & 0.228 & 1.000 & 6 & 20 & 2.54 \\
Claude API & claude-haiku-5-5 & mid & 40 & 10 & 0.052 & 0.223 & 0.479 & 0.998 & 5 & 40 & 3.66 \\
Claude API & claude-haiku-5-5 & long & 10 & 10 & 0.218 & 0.222 & 0.516 & 1.000 & 5 & 10 & 3.27 \\
\addlinespace
Bedrock & claude-haiku-5-5 & short & 20 & 10 & 0.070 & 0.143 & 0.190 & 0.997 & 4 & 20 & 3.31 \\
Bedrock & claude-haiku-5-5 & mid & 40 & 10 & 0.084 & 0.272 & 0.528 & 0.997 & 10 & 40 & 4.53 \\
Bedrock & claude-haiku-5-5 & long & 10 & 10 & 0.191 & 0.211 & 0.482 & 1.000 & 5 & 10 & 7.0 \\
\midrule
OpenAI API & gpt-6-astra & short & 20 & 10 & 0.042 & 0.041 & 0.043 & 0.969 & 2 & 20 & 3.0 \\
OpenAI API & gpt-6-astra & mid & 40 & 10 & 0.066 & 0.044 & 0.214 & 0.989 & 6 & 40 & 4.32 \\
OpenAI API & gpt-6-astra & long & 10 & 10 & 0.036 & 0.036 & 0.160 & 1.000 & 1 & 10 & 5.68 \\
\addlinespace
Azure OpenAI & gpt-6-astra & short & 20 & 10 & 0.036 & 0.033 & 0.036 & 0.947 & 2 & 20 & 2.9 \\
Azure OpenAI & gpt-6-astra & mid & 40 & 10 & 0.089 & 0.056 & 0.244 & 0.991 & 9 & 40 & 4.62 \\
Azure OpenAI & gpt-6-astra & long & 10 & 10 & 0.000 & 0.000 & 0.131 & 0.998 & 0 & 10 & 5.44 \\
\addlinespace
OpenAI API & gpt-6-luna & short & 20 & 10 & 0.030 & 0.000 & 0.000 & 0.999 & 3 & 20 & 2.3 \\
OpenAI API & gpt-6-luna & mid & 40 & 10 & 0.142 & 0.141 & 0.359 & 0.997 & 13 & 40 & 3.58 \\
OpenAI API & gpt-6-luna & long & 10 & 10 & 0.138 & 0.076 & 0.267 & 1.000 & 6 & 10 & 4.01 \\
\addlinespace
Azure OpenAI & gpt-6-luna & short & 20 & 10 & 0.030 & 0.020 & 0.020 & 1.000 & 3 & 20 & 2.36 \\
Azure OpenAI & gpt-6-luna & mid & 40 & 10 & 0.192 & 0.173 & 0.398 & 0.994 & 21 & 40 & 4.03 \\
Azure OpenAI & gpt-6-luna & long & 10 & 10 & 0.060 & 0.080 & 0.264 & 1.000 & 3 & 10 & 4.93 \\
\bottomrule
\end{tabular}
\end{adjustbox}
\end{sidewaystable*}

\begin{sidewaystable*}
\centering
\caption{Experiment 1 routes. ``Settings'' lists the determinism-related settings the route accepted; no reasoning or thinking parameter was sent, so reasoning ran at each provider's default on every route.}
\label{tab:e1routes}
\small
\begin{adjustbox}{max width=\linewidth}
\begin{tabular}{lllll}
\toprule
Route & Model (version returned) & Settings accepted & Region & Dates (2026, JST) \\
\midrule
OpenAI API & gpt-6-astra, gpt-6-luna & seed 0, JSON schema (temperature rejected) & default & Oct 4--5 \\
Azure OpenAI & gpt-6-astra (2026-09-03), gpt-6-luna (2026-09-22) & seed 0, JSON schema (temperature rejected) & Data Zone US, East US 2 & Oct 7--8 \\
Claude API & claude-opus-5-5 & JSON schema (temperature rejected) & \texttt{inference\_geo: us} & Oct 4--5 \\
Claude API & claude-haiku-4-5 (20251001) & temperature 0, JSON schema & default (geo not supported) & Oct 4--5 \\
Amazon Bedrock & claude-haiku-4-5 (20251001) & temperature 0, JSON schema & US inference profile & Oct 4--5 \\
Claude API & claude-haiku-5-5 & JSON schema (temperature rejected) & \texttt{inference\_geo: us} & Oct 12--13 \\
Amazon Bedrock & claude-opus-5-5, claude-haiku-5-5 & JSON schema sent$^\ast$ (temperature rejected) & US inference profile & Oct 12--13 \\
Gemini API & gemini-3.8-flash & seed 0, JSON schema (temperature 0 sent; ignored$^\dagger$) & default & Oct 4--5 \\
Agent Platform & gemini-3.8-flash & seed 0, JSON schema (temperature 0 sent; ignored$^\dagger$) & \texttt{us}, cache disabled & Oct 4--5 \\
Gemini API & gemini-3.6-flash & seed 0, JSON schema (temperature 0 sent; ignored$^\dagger$) & default & Oct 13--14 \\
Agent Platform & gemini-3.6-flash & seed 0, JSON schema (temperature 0 sent; ignored$^\dagger$) & \texttt{us}, cache disabled & Oct 13--14 \\
Ours (self-hosted) & gemma-4-31b & temperature 0, seed 0, JSON schema & us-central1 & Oct 5--6 \\
\bottomrule
\end{tabular}
\end{adjustbox}

\smallskip
\footnotesize $^\ast$ AWS documents structured outputs as unsupported for these models on Bedrock \citep{aws_opus55_card}; the request field was not rejected and every output parsed as JSON, but enforcement of the schema was not confirmed. $^\dagger$ Google documents temperature, top-p and top-k as ignored from Gemini 3.6 Flash onward \citep{google_gemini36}.
\end{sidewaystable*}

\begin{table*}[t]
\centering
\caption{Experiment 1, day 1 (70 prompts $\times$ 5 runs): calls, tokens returned by the provider, median latency, and cost estimated from those tokens at list prices; cache-read discounts and cache-write charges are both excluded, and the figures are estimates, not billed amounts. For our system, the cost is GPU time at the Spot price paid (\$2.25 per hour; about \$9.9 at the on-demand rate of \$5.85). Rows are in the order of Table~\ref{tab:e1}.}
\label{tab:e1cost}
\small
\begin{adjustbox}{max width=\linewidth}
\begin{tabular}{llrrrrr}
\toprule
Route & Model & Calls & Input tokens (M) & Output tokens (k) & Median s & Cost (USD) \\
\midrule
Ours (self-hosted) & gemma-4-31b & 350 & 10.5 & 45 & 8.5 & $\approx$3.8 \\
\midrule
Gemini API & gemini-3.8-flash & 350 & 10.5 & 369 & 3.6 & 9.26 \\
Agent Platform & gemini-3.8-flash & 350 & 10.6 & 383 & 6.1 & 10.30 \\
Gemini API$^\S$ & gemini-3.6-flash & 350 & 10.5 & 356 & 5.6 & 9.21 \\
Agent Platform$^\S$ & gemini-3.6-flash & 350 & 10.6 & 341 & 5.1 & 10.12 \\
\midrule
Claude API & claude-haiku-4-5 & 350 & 11.0 & 72 & 3.0 & 11.37 \\
Amazon Bedrock & claude-haiku-4-5 & 350 & 11.0 & 68 & 3.7 & 12.49 \\
Claude API & claude-opus-5-5 & 350 & 15.6 & 149 & 6.2 & 71.92 \\
Amazon Bedrock$^\ddagger$ & claude-opus-5-5 & 350 & 15.6 & 149 & 10.0 & 71.91 \\
Claude API$^\ddagger$ & claude-haiku-5-5 & 350 & 15.6 & 188 & 3.3 & 6.25 \\
Amazon Bedrock$^\ddagger$ & claude-haiku-5-5 & 350 & 15.6 & 187 & 4.7 & 6.25 \\
\midrule
OpenAI API & gpt-6-astra & 350 & 9.7 & 55 & 4.0 & 99.95 \\
Azure OpenAI$^\dagger$ & gpt-6-astra & 350 & 9.7 & 59 & 4.6 & 110.19 \\
OpenAI API & gpt-6-luna & 350 & 9.7 & 218 & 3.1 & 1.08 \\
Azure OpenAI$^\dagger$ & gpt-6-luna & 350 & 9.7 & 329 & 3.8 & 1.25 \\
\bottomrule
\end{tabular}
\end{adjustbox}

\smallskip
\footnotesize $^\dagger$ Day 1 is 7 October; Data Zone Standard prices. $^\ddagger$ Day 1 is 12 October. $^\S$ Day 1 is 13 October. Prices are the providers' list prices read on 14 October 2026: the Gemini 3 Flash rows at the introductory rate valid to 31 December 2026, the Agent Platform rows at its non-global (US) rate; the Claude API claude-haiku-4-5 row at the global rate, since that model does not support US-only inference; the Claude 5.5 rows and the Bedrock claude-haiku-4-5 row at the US regional rate. Output tokens include reasoning tokens where the provider reports them (for gpt-6-luna, 185 of 218 thousand through the OpenAI API and 296 of 329 thousand through Azure); reasoning tokens were reported on every call of the Gemini deployments and of gpt-6-luna, on 435 of 700 calls of gpt-6-astra and 503 of 700 of claude-opus-5-5 through their own APIs, on 237 of 350 of gpt-6-astra through Azure on day 1, on 481 of 700 of claude-opus-5-5 through Bedrock, on 612 and 621 of 700 of claude-haiku-5-5 through the Claude API and Bedrock, and on none of claude-haiku-4-5. A full run of the 776 pipeline assessments took 20.3 minutes of one GPU, about \$0.76 at the Spot price.
\end{table*}

\input{supp51}
\input{supp52}

\clearpage
\begin{table*}[t]
\centering
\caption{Experiment 1 across days (runs 1--5 on day 1, 6--10 on day 2, all tiers). Columns 3--5: the 70 prompts divided by how the label behaved over the ten runs---the same label in all ten; a label that varied within a day; or the same label in all five runs of each day but a different label on the two days. Columns 6--7: probability that two runs differ (all tiers), for pairs within day 1 / within day 2 / across days; the day-1 label values repeat the last column of Table~\ref{tab:e1} for comparison. Rows are in the order of Table~\ref{tab:e1}.}
\label{tab:e1days}
\small
\begin{adjustbox}{max width=\linewidth}
\begin{tabular}{llccccc}
\toprule
& & \multicolumn{3}{c}{Prompts by label behaviour (of 70)} & \multicolumn{2}{c}{$P$(two runs differ): day 1 / day 2 / across} \\
\cmidrule(lr){3-5}\cmidrule(lr){6-7}
Route & Model & \shortstack{Same in\\all 10 runs} & \shortstack{Varied\\within a day} & \shortstack{Changed only\\between days} & Label & Output \\
\midrule
Ours (self-hosted) & gemma-4-31b & 70 & 0 & 0 & 0.000 / 0.000 / 0.000 & 0.000 / 0.000 / 0.000 \\
\midrule
Gemini API & gemini-3.8-flash & 69 & 0 & 1 & 0.000 / 0.000 / 0.014 & 0.091 / 0.071 / 0.183 \\
Agent Platform & gemini-3.8-flash & 68 & 0 & 2 & 0.000 / 0.000 / 0.029 & 0.074 / 0.011 / 0.519 \\
Gemini API$^\S$ & gemini-3.6-flash & 70 & 0 & 0 & 0.000 / 0.000 / 0.000 & 0.463 / 0.443 / 0.407 \\
Agent Platform$^\S$ & gemini-3.6-flash & 68 & 2 & 0 & 0.009 / 0.017 / 0.013 & 0.484 / 0.497 / 0.489 \\
\midrule
Claude API & claude-haiku-4-5 & 66 & 4 & 0 & 0.034 / 0.023 / 0.026 & 0.319 / 0.330 / 0.361 \\
Amazon Bedrock & claude-haiku-4-5 & 70 & 0 & 0 & 0.000 / 0.000 / 0.000 & 0.011 / 0.023 / 0.015 \\
Claude API & claude-opus-5-5 & 56 & 14 & 0 & 0.081 / 0.086 / 0.077 & 1.000 / 0.999 / 1.000 \\
Amazon Bedrock$^\ddagger$ & claude-opus-5-5 & 56 & 14 & 0 & 0.080 / 0.069 / 0.078 & 0.999 / 1.000 / 0.999 \\
Claude API$^\ddagger$ & claude-haiku-5-5 & 54 & 16 & 0 & 0.091 / 0.083 / 0.094 & 1.000 / 1.000 / 0.998 \\
Amazon Bedrock$^\ddagger$ & claude-haiku-5-5 & 51 & 19 & 0 & 0.104 / 0.071 / 0.101 & 0.996 / 0.999 / 0.997 \\
\midrule
OpenAI API & gpt-6-astra & 61 & 9 & 0 & 0.046 / 0.053 / 0.059 & 0.981 / 0.977 / 0.989 \\
Azure OpenAI$^\dagger$ & gpt-6-astra & 59 & 11 & 0 & 0.077 / 0.043 / 0.062 & 0.979 / 0.974 / 0.981 \\
OpenAI API & gpt-6-luna & 48 & 22 & 0 & 0.103 / 0.123 / 0.107 & 0.999 / 0.997 / 0.998 \\
Azure OpenAI$^\dagger$ & gpt-6-luna & 43 & 27 & 0 & 0.137 / 0.114 / 0.128 & 0.994 / 0.999 / 0.997 \\
\bottomrule
\end{tabular}
\end{adjustbox}

\smallskip
\footnotesize $^\dagger$ 7--8 October; $^\ddagger$ 12--13 October; $^\S$ 13--14 October; the other commercial routes 4--5 October.
\end{table*}
\begin{table*}[t]
\centering
\caption{Retrieval baselines. Queries (of 33, one per distinct question, scored over all 4{,}977 chunks) whose top-6 results were identical in order to the first index of the same kind: hnswlib (approximate HNSW), Elasticsearch approximate $k$-nearest-neighbour search (ES kNN) and Elasticsearch BM25. ``Ours'' is the pipeline's exact, scoped retrieval, which matched in every case.}
\label{tab:baselines}
\begin{adjustbox}{max width=\linewidth}
\begin{tabular}{lcccc}
\toprule
Operation & hnswlib & ES kNN & ES BM25 & Ours \\
\midrule
Rebuild, same order, 1 thread / 1 bulk writer & 33 & 30 & 33 & 33 \\
Reverse order & 5 & 26 & 33 & 33 \\
Hash order & 15 & 27 & 33 & 33 \\
Parallel insertion (16 threads / 4 bulk writers) & 24--29 & 27--30 & 33 & 33 \\
Add then delete & 33 & 30 & 32$^\ast$ & 33 \\
Add then delete, force-merged & --- & 17 & 32$^\ast$ & 33 \\
Three shards instead of one & --- & 27 & 3$^\ast$ & 33 \\
\bottomrule
\end{tabular}
\end{adjustbox}

\smallskip
\footnotesize $^\ast$ BM25 scores changed for all 33 queries.
\end{table*}
\begin{table*}[p]
\centering\small
\caption{The pipeline's components, parameters and rules (Experiment 2), as recorded in the manifests. All hashes are SHA-256 over raw bytes, a model's raw output, or the canonical JSON (sorted keys) of a structured value.}
\label{tab:params}
\begin{tabularx}{\linewidth}{lX}
\toprule
Stage & Component, parameters and rule \\
\midrule
Extraction & pdf.js text layer; a page is read by OCR when its text layer, with white space removed, has fewer than 200 characters; the page is rendered at 150 dpi and decoded greedily (temperature 0, seed 0, at most 6{,}000 tokens) by \texttt{allenai/olmOCR-2-7B-1025}, with \texttt{zai-org/GLM-OCR} used only when olmOCR-2 stops at the token limit; both served by vLLM 0.29.0 with batch invariance. Hashes: original (file bytes); provenance (original hash, extractor manifest, each page's method); version (text, metadata, provenance hash). \\
Chunking & \texttt{splitText} v1: chunks of at most 1{,}400 characters overlapping by 120, cut at the last paragraph or sentence boundary in the final 40\% of the window when there is one. Hash: the chunk-set hash over the chunk strings. \\
Embedding & \texttt{Qwen/Qwen3-Embedding-8B}, bfloat16, PyTorch deterministic algorithms (\texttt{CUBLAS\_WORKSPACE\_CONFIG=:4096:8}), one text at a time on the GPU; document vectors unit-normalised and stored as float32; the question embedded at run time with the instruction prefix ``Instruct: Given a query, retrieve passages that answer or support it''. Hashes: embedding-set identifier (version hash, chunker, embedder manifest); vector hash over the float32 vectors in sequence. \\
Retrieval & Tokenizer \texttt{simple} v1: lower-case, NFKC, maximal runs of letters and digits, no stop words or stemming. BM25 with $k_1 = 1.2$, $b = 0.75$, statistics over the chunks in scope (the device's documents). Cosine similarity as the dot product of unit vectors. Each ranking truncated to its top 30, min--max normalised, combined with weights 0.5 and 0.5; ties broken by document identifier, then chunk index; top six kept. Hash: query, parameters and scored hits. \\
Judgment & \texttt{RedHatAI/gemma-4-31B-it-FP8-dynamic} on vLLM 0.29.0, \texttt{VLLM\_BATCH\_INVARIANT=1}, \texttt{TRITON\_ATTN}, enforce-eager, chunked prefill with \texttt{max-num-batched-tokens} 16{,}600, \texttt{max-num-seqs} 16, FP8 KV cache, xgrammar structured output without white space, temperature 0, seed 0, at most 1{,}024 output tokens. Hashes: prompt (the chat messages); raw output (the output string). \\
Verification and record & A quotation and the passage it cites are NFKC-normalised, all quotation marks unified and all white space removed; each part of the quotation around an ellipsis must be an exact substring of the passage; failed quotations are recorded as rejected and removed, and the label is downgraded to \emph{Cannot determine} only when no quotation remains. The ledger chains entries by hashing each with the previous entry's hash. \\
\bottomrule
\end{tabularx}
\end{table*}

\clearpage
\subsection*{Supplementary Note 1: the prompts of Experiment 1}
 The system prompt (SHA-256 \texttt{1181bf6aa31300b525d1fc20e6d24e0c}\allowbreak\texttt{8e00d27c6b69414d47d6e6db46fcc0d9} over its UTF-8 bytes, with straight quotation marks and the line breaks shown) is reproduced verbatim:
\begin{quote}\small You review FDA 510(k) premarket documents for a blood glucose monitoring system and decide whether the documents show that the device meets each requirement taken from FDA guidance. Use only the documents provided.

Decide one requirement at a time. Choose exactly one label:\\
- Satisfied: The documents show the requirement is met.\\
- Not satisfied: The documents show the requirement is not met.\\
- Cannot determine: The documents do not address the requirement.

Answer with JSON only, in this form: \{\texttt{"}label\texttt{"}: \texttt{"}\textless{}one of the labels\textgreater{}\texttt{"}, \texttt{"}comment\texttt{"}: \texttt{"}\textless{}two or three sentences explaining the decision\textgreater{}\texttt{"}, \texttt{"}evidence\texttt{"}: [\{\texttt{"}document\texttt{"}: \textless{}document number\textgreater{}, \texttt{"}quote\texttt{"}: \texttt{"}\textless{}a passage copied verbatim from that document\textgreater{}\texttt{"}\}]\}. Quote passages exactly as they appear in the documents. If the label is Cannot determine, return an empty evidence list.\end{quote}
The user prompt is the numbered documents, each introduced by a line ``\texttt{=== Document }$n$\texttt{: }\textit{title}\texttt{ ===}'', followed by ``\texttt{Requirement: }\textit{item}'' and ``\texttt{Answer with JSON only.}''. The item ends with ``\texttt{(FDA guidance section: }\textit{heading}\texttt{)}'', where the heading is in fact that of the device's decision-summary section matched to the item, not a section of the guidance; the same text is used in both experiments, so this misnomer does not affect the comparisons. The JSON schema sent where accepted requires an object with exactly the keys \texttt{label} (one of the three labels), \texttt{comment} (string) and \texttt{evidence} (an array of objects with an integer \texttt{document} and a string \texttt{quote}) and no other properties. Experiment~2 uses the same label definitions. All 70 prompts and their SHA-256 are in the supplementary material (\texttt{prompts\_index.jsonl}); full texts are reconstructible from the public documents.

\subsection*{Supplementary Note 2: changes during the study}
 The measures, operations and conditions were fixed before the runs, and every change was recorded when it was made. The positive-control runs (``Ours''), the concurrency test, the extension of day~2 to all tiers and the infusion-pump runs were added during the study and are exploratory. The changes were as follows.
\begin{description}[leftmargin=*,itemsep=2pt,font=\normalfont\itshape]
\item[Experiment 1.] (i) Device--item pairs whose 510(k) summary had under 1{,}000 characters of text (two PDF portfolios) were skipped in favour of the next pair. (ii) On Amazon Bedrock, \texttt{claude-haiku-4-5} was measured on 4--5 October in place of \texttt{claude-opus-5-5}, so that both Claude routes carried the same model; \texttt{claude-opus-5-5} through Bedrock and \texttt{claude-haiku-5-5} through both Claude routes were added on 12--13 October, with \texttt{claude-opus-5-5} through the Claude API sent again on those days for the route comparison (its 4--5 October measurement remains the one reported in Tables~\ref{tab:e1} and~\ref{tab:e1days}). An earlier attempt on 8 October was interrupted after 105 Bedrock calls and is not used. (iii) The Claude API credit balance ran out for 16 seconds; the 78 failed calls were resent. (iv) The second day was extended from mid prompts to all tiers, and the self-hosted judgment stage was added with the same prompts and schedule. (v) Azure OpenAI was measured on 7--8 October through an existing subscription, with the OpenAI API's short prompts sent on the same days for the route comparison. (vi) \texttt{gemini-3.5-flash} through the Gemini API had been planned for the mid prompts only; short and long prompts were added on 7--8 October so that every deployment has the same 70 prompts. On 13 October, the identifier \texttt{gemini-3.5-flash} returned the version \texttt{gemini-3.6-flash} through the Gemini API (it had returned \texttt{gemini-3.5-flash} on 4--8 October), while Agent Platform still served \texttt{gemini-3.5-flash} under it, so the two routes could no longer be compared on the same days under that identifier; we therefore measured \texttt{gemini-3.6-flash} through both Google routes on 13--14 October and report it in place of the earlier measurement, which remains in the supplementary ledger but appears in no table; Google's release notes of 8 October 2026 record the deprecation and the automatic routing \citep{google_changelog}. (vii) The cost cap was raised in steps from \$480 to \$1{,}300. (viii) The count of label changes between days was defined after the data were collected, as in Measures and statistics; a count of prompts with any differing pair of runs across days, used in drafts, also counted variation within a day. The within- and across-day probabilities that two runs differ, defined before the runs, are reported unchanged. (ix) The reduced-reasoning and retrieved-passage measurements (Supplementary Tables~\ref{tab:e4reason} and~\ref{tab:e3prompts}) were added on 14 October after internal review, pre-registered before sending, and run on 14--15 October; their first launch sent no calls because the runner skips completed (route, model, prompt, run) tuples, so the reduced-reasoning runs use run numbers 31--40, mapped to 1--10 in the tables.
\item[Concurrency test.] (i) The test was added after Experiment~1 had begun. (ii) The second model was changed before measurement from Qwen3.8-27B, which the serving software refused to run in batch-invariant mode, to Qwen3-32B on the 60 short and mid prompts, with thinking disabled and the attention kernel fixed; Qwen3.8-27B was measured without batch invariance only.
\item[Experiment 2.] (i) After the dry runs, FDA guidance was removed from the corpus, because the items are guidance sentences and retrieval returned the guidance itself, and the item count was corrected to 776 unique device--item pairs. (ii) Ingestion with 16 concurrent documents stalled, because page rendering in the PDF library did not return when documents were held open across long OCR waits; the extractor was changed to render all OCR pages before calling OCR, extracted text is unchanged, and the rerun matched the reference run produced before the fix. (iii) The other-GPU test was planned on an A100 80GB, which could not run the reference configuration (in vLLM 0.29.0's Triton attention backend, used with batch invariance, an FP8 KV cache requires compute capability 8.9, which the A100 lacks, and the FP8 weight conversion kernel failed to compile); it was run on an H100 80GB with the same settings and the memory share raised from 0.62 to 0.74 to give the same absolute memory. (iv) A run on a later day and the infusion-pump runs were added. (v) T8 and T9 were added on 10 October, after reading Vosti \citep{qin2026vosti}; the smaller chunk budget was changed from 2{,}048 to 4{,}096 tokens before measurement because the engine rejected 2{,}048 for this multimodal model. (vi) After internal review, on 14--15 October, we added a second H100 run on a fresh machine created from a disk image of the reference machine, the repeats of the 4{,}096 budget at 1, 8 and 16 concurrent items, the server-restart and interrupted-ingestion runs, the OCR and embedding ablations and the hnswlib and soft-delete tests (Methods); the OCR ablation was stopped by its two-hour limit after 336 of 337 documents, which are the documents compared.
\end{description}

\end{document}

%% file: fig2.tex
\begin{tikzpicture}
\begin{axis}[
  width=0.66\textwidth, height=6.6cm,
  xmin=-1.5, xmax=73, xtick={0,10,...,70}, xlabel={Prompts, of 70, on which the field differed in at least one of 5 runs},
  ytick={1,...,15}, y dir=reverse, ymin=0.3, ymax=15.7,
  yticklabels={Ours (self-hosted) / gemma-4-31b, Gemini API / gemini-3.8-flash, Agent Platform / gemini-3.8-flash, Gemini API / gemini-3.6-flash, Agent Platform / gemini-3.6-flash, Claude API / claude-haiku-4-5, Amazon Bedrock / claude-haiku-4-5, Claude API / claude-opus-5-5, Amazon Bedrock / claude-opus-5-5, Claude API / claude-haiku-5-5, Amazon Bedrock / claude-haiku-5-5, OpenAI API / gpt-6-astra, Azure OpenAI / gpt-6-astra, OpenAI API / gpt-6-luna, Azure OpenAI / gpt-6-luna},
  yticklabel style={font=\footnotesize}, xticklabel style={font=\footnotesize}, xlabel style={font=\footnotesize}, ytick style={draw=none},
  xmajorgrids, grid style={gray!20}, axis line style={gray!60},
  legend style={at={(0.5,1.02)}, anchor=south, legend columns=3, font=\footnotesize, draw=none},
  clip=false,
]
\addplot[gray!60, line width=1.2pt, forget plot] coordinates {(0,1) (0,1)};
\addplot[gray!60, line width=1.2pt, forget plot] coordinates {(0,2) (14,2)};
\addplot[gray!60, line width=1.2pt, forget plot] coordinates {(0,3) (11,3)};
\addplot[gray!60, line width=1.2pt, forget plot] coordinates {(0,4) (52,4)};
\addplot[gray!60, line width=1.2pt, forget plot] coordinates {(1,5) (61,5)};
\addplot[gray!60, line width=1.2pt, forget plot] coordinates {(4,6) (31,6)};
\addplot[gray!60, line width=1.2pt, forget plot] coordinates {(0,7) (2,7)};
\addplot[gray!60, line width=1.2pt, forget plot] coordinates {(11,8) (70,8)};
\addplot[gray!60, line width=1.2pt, forget plot] coordinates {(10,9) (70,9)};
\addplot[gray!60, line width=1.2pt, forget plot] coordinates {(13,10) (70,10)};
\addplot[gray!60, line width=1.2pt, forget plot] coordinates {(14,11) (70,11)};
\addplot[gray!60, line width=1.2pt, forget plot] coordinates {(6,12) (70,12)};
\addplot[gray!60, line width=1.2pt, forget plot] coordinates {(10,13) (70,13)};
\addplot[gray!60, line width=1.2pt, forget plot] coordinates {(14,14) (70,14)};
\addplot[gray!60, line width=1.2pt, forget plot] coordinates {(18,15) (70,15)};
\draw[gray!45] (axis cs:-1.5,1.5) -- (axis cs:73,1.5);
\draw[gray!45] (axis cs:-1.5,5.5) -- (axis cs:73,5.5);
\draw[gray!45] (axis cs:-1.5,11.5) -- (axis cs:73,11.5);
\addplot[only marks, mark=triangle*, mark size=2.6pt, red!75!black] coordinates {(0,1) (14,2) (11,3) (52,4) (61,5) (31,6) (2,7) (70,8) (70,9) (70,10) (70,11) (70,12) (70,13) (70,14) (70,15)};
\addlegendentry{Full output}
\addplot[only marks, mark=square*, mark size=2.0pt, blue!70!black] coordinates {(0,1) (0,2) (0,3) (3,4) (3,5) (11,6) (1,7) (23,8) (28,9) (42,10) (42,11) (11,12) (15,13) (20,14) (25,15)};
\addlegendentry{Cited passages}
\addplot[only marks, mark=*, mark size=2.2pt, black] coordinates {(0,1) (0,2) (0,3) (0,4) (1,5) (4,6) (0,7) (11,8) (10,9) (13,10) (14,11) (6,12) (10,13) (14,14) (18,15)};
\addlegendentry{Label}
\end{axis}
\end{tikzpicture}

%% file: supp51.tex
\begin{sidewaystable*}
\centering\footnotesize
\caption{Same-day comparison of the two routes of each model (Experiment 1). For each field, the number of the common prompts on which the field differed in at least one run through route 1 / through route 2, and the two-sided $p$ value of an exact McNemar test on the paired prompt-level indicators. ``Day 1'' uses the five runs of day 1; ``both days'' uses all ten runs. \texttt{claude-opus-5-5} is the 12--13 October measurement through both routes; the OpenAI models are the 20 short prompts of 7--8 October. The test treats prompts as independent.}
\label{tab:routes}
\begin{tabular}{llcllllll}
\toprule
Model & Routes 1 / 2 & Prompts & Label & Files & Passages & Quotes & Comment & Output \\
\midrule
\multicolumn{9}{l}{\emph{Day 1}} \\
\texttt{claude-haiku-4-5} & Claude API / Amazon Bedrock & 70 & 4 / 0 (0.125) & 5 / 0 (0.062) & 11 / 1 (0.006) & 13 / 1 (0.002) & 30 / 2 ($<$0.001) & 31 / 2 ($<$0.001) \\
\texttt{claude-opus-5-5} & Claude API / Amazon Bedrock & 70 & 12 / 10 (0.754) & 9 / 13 (0.289) & 29 / 28 (1.000) & 38 / 38 (1.000) & 70 / 70 (1.000) & 70 / 70 (1.000) \\
\texttt{claude-haiku-5-5} & Claude API / Amazon Bedrock & 70 & 13 / 14 (1.000) & 32 / 36 (0.541) & 42 / 42 (1.000) & 44 / 46 (0.754) & 70 / 70 (1.000) & 70 / 70 (1.000) \\
\texttt{gpt-6-astra} & OpenAI API / Azure OpenAI & 20 & 1 / 1 (1.000) & 1 / 1 (1.000) & 1 / 1 (1.000) & 1 / 1 (1.000) & 20 / 20 (1.000) & 20 / 20 (1.000) \\
\texttt{gpt-6-luna} & OpenAI API / Azure OpenAI & 20 & 2 / 1 (1.000) & 2 / 1 (1.000) & 2 / 1 (1.000) & 2 / 1 (1.000) & 20 / 20 (1.000) & 20 / 20 (1.000) \\
\texttt{gemini-3.8-flash} & Gemini API / Agent Platform & 70 & 0 / 0 (1.000) & 0 / 0 (1.000) & 0 / 0 (1.000) & 0 / 0 (1.000) & 14 / 11 (0.581) & 14 / 11 (0.581) \\
\texttt{gemini-3.6-flash} & Gemini API / Agent Platform & 70 & 0 / 1 (1.000) & 1 / 1 (1.000) & 3 / 3 (1.000) & 7 / 6 (1.000) & 52 / 61 (0.108) & 52 / 61 (0.108) \\
\midrule
\multicolumn{9}{l}{\emph{Both days}} \\
\texttt{claude-haiku-4-5} & Claude API / Amazon Bedrock & 70 & 4 / 0 (0.125) & 7 / 0 (0.016) & 15 / 1 (0.001) & 17 / 1 ($<$0.001) & 38 / 3 ($<$0.001) & 38 / 3 ($<$0.001) \\
\texttt{claude-opus-5-5} & Claude API / Amazon Bedrock & 70 & 17 / 14 (0.375) & 12 / 15 (0.508) & 33 / 31 (0.774) & 40 / 40 (1.000) & 70 / 70 (1.000) & 70 / 70 (1.000) \\
\texttt{claude-haiku-5-5} & Claude API / Amazon Bedrock & 70 & 16 / 19 (0.508) & 39 / 45 (0.210) & 44 / 50 (0.070) & 46 / 50 (0.219) & 70 / 70 (1.000) & 70 / 70 (1.000) \\
\texttt{gpt-6-astra} & OpenAI API / Azure OpenAI & 20 & 1 / 2 (1.000) & 1 / 2 (1.000) & 1 / 2 (1.000) & 1 / 2 (1.000) & 20 / 20 (1.000) & 20 / 20 (1.000) \\
\texttt{gpt-6-luna} & OpenAI API / Azure OpenAI & 20 & 2 / 3 (1.000) & 2 / 2 (1.000) & 2 / 2 (1.000) & 2 / 2 (1.000) & 20 / 20 (1.000) & 20 / 20 (1.000) \\
\texttt{gemini-3.8-flash} & Gemini API / Agent Platform & 70 & 1 / 2 (1.000) & 1 / 2 (1.000) & 1 / 2 (1.000) & 1 / 6 (0.062) & 22 / 41 (0.003) & 22 / 41 (0.003) \\
\texttt{gemini-3.6-flash} & Gemini API / Agent Platform & 70 & 0 / 2 (0.500) & 1 / 2 (1.000) & 3 / 4 (1.000) & 7 / 8 (1.000) & 54 / 70 ($<$0.001) & 54 / 70 ($<$0.001) \\
\bottomrule
\end{tabular}
\end{sidewaystable*}

\begin{table*}[p]
\centering\small
\caption{Experiment 1 by cache status (deployments whose responses report cached input tokens). Pairs of runs of the same prompt are grouped by whether neither call, one call or both calls reported a cache hit; for each group, the number of pairs and the share that differed in label / in output. Through the OpenAI API and Azure OpenAI, 562 of 700 calls on each deployment reported cached tokens. Agent Platform ran with implicit caching disabled and the Claude routes reported no cache reads, so they are not listed. Within: pairs on the same day; across: pairs on different days.}
\label{tab:cache}
\begin{adjustbox}{max width=\linewidth}
\begin{tabular}{llllll}
\toprule
Route & Model & Scope & Neither cached & One cached & Both cached \\
\midrule
Azure OpenAI & \texttt{gpt-6-astra} & within & --- & 552: 0.065 / 0.982 & 848: 0.057 / 0.973 \\
Azure OpenAI & \texttt{gpt-6-astra} & across & 68: 0.088 / 0.985 & 554: 0.061 / 0.982 & 1128: 0.060 / 0.980 \\
Azure OpenAI & \texttt{gpt-6-luna} & within & --- & 552: 0.147 / 0.998 & 848: 0.112 / 0.995 \\
Azure OpenAI & \texttt{gpt-6-luna} & across & 68: 0.162 / 1.000 & 554: 0.152 / 0.996 & 1128: 0.114 / 0.997 \\
Gemini API & \texttt{gemini-3.6-flash} & within & 599: 0.000 / 0.002 & 498: 0.000 / 0.998 & 303: 0.000 / 0.449 \\
Gemini API & \texttt{gemini-3.6-flash} & across & 790: 0.000 / 0.011 & 540: 0.000 / 0.998 & 420: 0.000 / 0.390 \\
Gemini API & \texttt{gemini-3.8-flash} & within & 308: 0.000 / 0.036 & 516: 0.000 / 0.155 & 576: 0.000 / 0.040 \\
Gemini API & \texttt{gemini-3.8-flash} & across & 417: 0.012 / 0.094 & 581: 0.034 / 0.305 & 752: 0.000 / 0.140 \\
OpenAI API & \texttt{gpt-6-astra} & within & --- & 552: 0.040 / 0.975 & 848: 0.055 / 0.982 \\
OpenAI API & \texttt{gpt-6-astra} & across & 68: 0.059 / 0.971 & 554: 0.058 / 0.986 & 1128: 0.059 / 0.992 \\
OpenAI API & \texttt{gpt-6-luna} & within & --- & 552: 0.101 / 0.998 & 848: 0.120 / 0.998 \\
OpenAI API & \texttt{gpt-6-luna} & across & 68: 0.088 / 1.000 & 554: 0.101 / 0.996 & 1128: 0.111 / 0.999 \\
\bottomrule
\end{tabular}
\end{adjustbox}
\end{table*}

\begin{table*}[p]
\centering\small
\caption{Experiment 1, day 1: prompts whose label was the same in all five runs (``stable''), and among them the prompts whose cited passages, or whose exact quotation strings, still differed; prompts whose label changed, and among them those on which one of the labels was \emph{Cannot determine}; and, over both days, the quotations returned and the share located verbatim (after normalising white space and case) in the cited document. Rows are in the order of Table~\ref{tab:e1}.}
\label{tab:stable}
\begin{adjustbox}{max width=\linewidth}
\begin{tabular}{llrrrrrrr}
\toprule
Route & Model & Label stable & Passages differ & Quotes differ & Label changed & Involving CD & Quotations & Located \\
\midrule
Ours (self-hosted) & \texttt{gemma-4-31b} & 70 & 0 & 0 & 0 & 0 & --- & --- \\
Gemini API & \texttt{gemini-3.8-flash} & 70 & 0 & 0 & 0 & 0 & 395 & 94.9\% \\
Agent Platform & \texttt{gemini-3.8-flash} & 70 & 0 & 0 & 0 & 0 & 390 & 94.9\% \\
Gemini API & \texttt{gemini-3.6-flash} & 70 & 3 & 7 & 0 & 0 & 376 & 94.7\% \\
Agent Platform & \texttt{gemini-3.6-flash} & 69 & 2 & 5 & 1 & 1 & 399 & 95.0\% \\
Claude API & \texttt{claude-haiku-4-5} & 66 & 8 & 10 & 4 & 4 & 683 & 90.5\% \\
Amazon Bedrock & \texttt{claude-haiku-4-5} & 70 & 1 & 1 & 0 & 0 & 630 & 87.8\% \\
Claude API & \texttt{claude-opus-5-5} & 59 & 13 & 26 & 11 & 8 & 859 & 99.1\% \\
Amazon Bedrock & \texttt{claude-opus-5-5} & 60 & 18 & 28 & 10 & 8 & 857 & 99.2\% \\
Claude API & \texttt{claude-haiku-5-5} & 57 & 30 & 31 & 13 & 13 & 600 & 98.5\% \\
Amazon Bedrock & \texttt{claude-haiku-5-5} & 56 & 29 & 32 & 14 & 14 & 564 & 96.8\% \\
OpenAI API & \texttt{gpt-6-astra} & 64 & 5 & 13 & 6 & 5 & 443 & 99.5\% \\
Azure OpenAI & \texttt{gpt-6-astra} & 60 & 7 & 12 & 10 & 3 & 435 & 99.1\% \\
OpenAI API & \texttt{gpt-6-luna} & 56 & 10 & 12 & 14 & 12 & 353 & 94.1\% \\
Azure OpenAI & \texttt{gpt-6-luna} & 52 & 9 & 14 & 18 & 13 & 348 & 93.7\% \\
\bottomrule
\end{tabular}
\end{adjustbox}
\end{table*}

\begin{table*}[p]
\centering\small
\caption{Concurrency test: the 70 prompts sent five times each from eight concurrent senders to one vLLM server, with batch-invariant inference (BI) enabled or disabled. Columns: prompts on which the field differed in at least one of the five concurrent runs, and prompts on which any concurrent output differed from the output of the sequential runs of Tables~\ref{tab:e1} and~\ref{tab:e1days} (Gemma only). Qwen3-32B: 57 of the 60 short and mid prompts fit its context of 40{,}960 tokens. Qwen3.8-27B could not be served with BI. Batch invariance lengthened the 350 concurrent Gemma calls from 55.9 to 65.9 minutes.}
\label{tab:concurrency}
\begin{adjustbox}{max width=\linewidth}
\begin{tabular}{llrrrrrrrr}
\toprule
Model & BI & Prompts & Label & Files & Passages & Quotes & Comment & Output & vs.\ sequential \\
\midrule
Gemma 4 31B (FP8) & on & 70 & 0 & 0 & 0 & 0 & 0 & 0 & 0 \\
Gemma 4 31B (FP8) & off & 70 & 5 & 6 & 8 & 14 & 58 & 58 & 64 \\
Qwen3-32B (BF16) & on & 57 & 0 & 0 & 0 & 0 & 0 & 0 & --- \\
Qwen3-32B (BF16) & off & 57 & 4 & 12 & 24 & 32 & 48 & 50 & --- \\
Qwen3.8-27B (FP8) & off & 70 & 12 & 15 & 18 & 20 & 69 & 69 & --- \\
\bottomrule
\end{tabular}
\end{adjustbox}
\end{table*}

\begin{table*}[p]
\centering\small
\caption{Experiment 1: calls and estimated cost by deployment and measurement day (list prices, from provider-reported tokens; Supplementary Table~\ref{tab:e1cost}), with the start of the first call and the end of the last call in Japan Standard Time. ``Calls'' counts calls that returned status 200; failed calls (78 on the Claude API on 4 October, 245 of the interrupted Bedrock attempt on 8 October) are not counted. Rows marked ``not used'' or ``not reported'' are in the deposited ledger but in no table; the self-hosted runs are not listed. The totals of the used rows are \$577 through 7 October, \$130 on 8 October, \$313 on 12--13 October and \$39 on 13--14 October.}
\label{tab:costdates}
\begin{adjustbox}{max width=\linewidth}
\begin{tabular}{llllrrl}
\toprule
Route & Model & Date (JST) & First call -- last call & Calls & Cost (USD) & Note \\
\midrule
Gemini API & \texttt{gemini-3.5-flash} & 04 Oct & 13:06--13:25 & 200 & 6.94 & gemini-3.5-flash, not reported \\
Gemini API & \texttt{gemini-3.8-flash} & 04 Oct & 13:06--13:34 & 350 & 9.26 &  \\
OpenAI API & \texttt{gpt-6-astra} & 04 Oct & 13:06--13:33 & 350 & 99.95 &  \\
Claude API & \texttt{claude-opus-5-5} & 04 Oct & 13:06--13:55 & 350 & 71.92 &  \\
OpenAI API & \texttt{gpt-6-luna} & 04 Oct & 13:06--13:35 & 350 & 1.08 &  \\
Agent Platform & \texttt{gemini-3.8-flash} & 04 Oct & 13:06--14:04 & 350 & 10.30 &  \\
Amazon Bedrock & \texttt{claude-haiku-4-5} & 04 Oct & 13:39--14:01 & 350 & 12.49 &  \\
Claude API & \texttt{claude-haiku-4-5} & 04 Oct & 13:39--13:58 & 350 & 11.37 &  \\
Amazon Bedrock & \texttt{claude-haiku-4-5} & 05 Oct & 10:26--13:01 & 350 & 12.49 &  \\
Gemini API & \texttt{gemini-3.8-flash} & 05 Oct & 10:26--13:02 & 350 & 9.19 &  \\
Gemini API & \texttt{gemini-3.5-flash} & 05 Oct & 10:26--10:45 & 200 & 6.92 & gemini-3.5-flash, not reported \\
Claude API & \texttt{claude-haiku-4-5} & 05 Oct & 10:26--12:58 & 350 & 11.37 &  \\
OpenAI API & \texttt{gpt-6-astra} & 05 Oct & 10:26--13:02 & 350 & 100.83 &  \\
OpenAI API & \texttt{gpt-6-luna} & 05 Oct & 10:26--13:08 & 350 & 1.09 &  \\
Claude API & \texttt{claude-opus-5-5} & 05 Oct & 10:26--13:05 & 350 & 71.90 &  \\
Agent Platform & \texttt{gemini-3.8-flash} & 05 Oct & 10:26--13:23 & 350 & 10.27 &  \\
Azure OpenAI & \texttt{gpt-6-astra} & 07 Oct & 12:13--12:54 & 350 & 110.19 &  \\
Azure OpenAI & \texttt{gpt-6-luna} & 07 Oct & 12:13--13:03 & 350 & 1.25 &  \\
OpenAI API & \texttt{gpt-6-luna} & 07 Oct & 12:13--12:17 & 100 & 0.07 & short prompts, route comparison \\
OpenAI API & \texttt{gpt-6-astra} & 07 Oct & 12:13--12:20 & 100 & 6.16 & short prompts, route comparison \\
Gemini API & \texttt{gemini-3.5-flash} & 07 Oct & 14:44--14:58 & 150 & 12.18 & gemini-3.5-flash, not reported \\
Gemini API & \texttt{gemini-3.5-flash} & 08 Oct & 07:58--08:12 & 150 & 12.18 & gemini-3.5-flash, not reported \\
Azure OpenAI & \texttt{gpt-6-luna} & 08 Oct & 07:58--08:45 & 350 & 1.21 &  \\
Azure OpenAI & \texttt{gpt-6-astra} & 08 Oct & 07:58--08:37 & 350 & 110.12 &  \\
OpenAI API & \texttt{gpt-6-luna} & 08 Oct & 07:58--08:03 & 100 & 0.07 & short prompts, route comparison \\
OpenAI API & \texttt{gpt-6-astra} & 08 Oct & 07:58--08:05 & 100 & 6.15 & short prompts, route comparison \\
Claude API & \texttt{claude-opus-5-5} & 08 Oct & 10:21--10:54 & 350 & 72.24 & interrupted attempt, not used \\
Amazon Bedrock & \texttt{claude-opus-5-5} & 08 Oct & 10:21--10:46 & 105 & 18.18 & interrupted attempt, not used \\
Claude API & \texttt{claude-opus-5-5} & 12 Oct & 09:12--09:46 & 350 & 71.94 & route comparison \\
Claude API & \texttt{claude-haiku-5-5} & 12 Oct & 09:12--09:32 & 350 & 6.25 &  \\
Amazon Bedrock & \texttt{claude-opus-5-5} & 12 Oct & 09:12--10:16 & 350 & 71.91 &  \\
Amazon Bedrock & \texttt{claude-haiku-5-5} & 12 Oct & 09:12--10:19 & 350 & 6.25 &  \\
Claude API & \texttt{claude-haiku-5-5} & 13 Oct & 00:08--00:29 & 350 & 6.25 &  \\
Claude API & \texttt{claude-opus-5-5} & 13 Oct & 00:08--00:42 & 350 & 71.98 & route comparison \\
Amazon Bedrock & \texttt{claude-opus-5-5} & 13 Oct & 00:08--01:06 & 350 & 71.86 &  \\
Amazon Bedrock & \texttt{claude-haiku-5-5} & 13 Oct & 00:08--01:04 & 350 & 6.25 &  \\
Gemini API & \texttt{gemini-3.6-flash} & 13 Oct & 13:01--13:39 & 350 & 9.21 &  \\
Agent Platform & \texttt{gemini-3.6-flash} & 13 Oct & 13:01--13:37 & 350 & 10.12 &  \\
Gemini API & \texttt{gemini-3.6-flash} & 14 Oct & 03:32--04:10 & 350 & 9.20 &  \\
Agent Platform & \texttt{gemini-3.6-flash} & 14 Oct & 03:32--04:08 & 350 & 10.10 &  \\
\bottomrule
\end{tabular}
\end{adjustbox}
\end{table*}

\begin{table*}[p]
\centering\scriptsize
\caption{The 12 assessment items of the infusion-pump corpus: of the 107 candidate sentences of Section~5 of FDA's infusion pump guidance, the sentence at position $\lfloor 107j/12 \rfloor$, $j = 0, \ldots, 11$ (Methods). Each item is sent as the sentence followed by ``(FDA guidance section: \ldots)'' naming its subsection. The reference run labelled 422 of the 516 pairs \emph{Cannot determine}, 84 \emph{Satisfied} and 2 \emph{Not satisfied}, with 8 unparsable outputs identical across the four runs and 86 assessments citing at least one quotation.}
\label{tab:frn}
\begin{tabularx}{\linewidth}{lX}
\toprule
Item & Sentence (subsection) \\
\midrule
frn-01 & For this reason, any newly developed device should have its own unique safety case. \emph{(5. Safety Assurance Case)} \\
frn-02 & The safety case for a specific infusion pump should incorporate the following: $\bullet$ The persons for whose use the device is represented or intended; and $\bullet$ The conditions of use for the device, including conditions of use prescribed, recommended, or suggested in the labeling or advertising of the device, and other intended conditions of use. \emph{(Safety Assurance Case: A. General Considerations for Safety Case Development)} \\
frn-03 & Individual pump designs, environmental conditions, and conditions of use for the device will dictate what hazards should be identified and mitigated. \emph{(Safety Assurance Case: B. Hazard Analysis)} \\
frn-04 & Your safety assurance case should address the appropriateness of any standard to support clearance of your infusion pump. \emph{(Safety Assurance Case: C. Performance Testing)} \\
frn-05 & For a pump that does not maintain a constant flow rate, test results should be used to generate a representative flow profile. \emph{(Safety Assurance Case: C. Performance Testing)} \\
frn-06 & The analysis should include a description of your system's reliability specification and the reliability activities completed to verify and validate that the specification has been met (e.g., design analysis, test plans, and test reports). \emph{(Safety Assurance Case: C. Performance Testing)} \\
frn-07 & Device manufacturers should consider appropriate levels of testing in accordance with the risks present in the intended use environments. \emph{(Safety Assurance Case: C. Performance Testing)} \\
frn-08 & When providing a list of unresolved anomalies, you should include the following information for each unresolved anomaly: $\bullet$ A description of the anomaly from a symptom point of view and how it is manifested. $\bullet$ The location in the code where the anomaly occurs. $\bullet$ A description of how to fix the anomalous code. $\bullet$ A search of the software source code for other possible instances of the anomaly. \emph{(Safety Assurance Case: C. Performance Testing)} \\
frn-09 & We recommend that if you consider the tests suggested in the referenced biocompatibility guidance document or material characterization testing discussed below to not be applicable or appropriate, you should provide adequate justification for omission of these tests. \emph{(Safety Assurance Case: C. Performance Testing)} \\
frn-10 & For the L\&E study, we recommend that you consider ISO 10993-18 Biological evaluation of medical devices --- Part 18: Chemical characterization of materials, 2005. $\bullet$ For device-related particulate evaluation, you should follow current USP $<$788$>$ Particulate Matter in Injections. \emph{(Safety Assurance Case: C. Performance Testing)} \\
frn-11 & Shelf Life If your particular infusion pump contains sterile components or materials that could degrade over time, we recommend that you include a shelf life on the packaging. \emph{(Safety Assurance Case: C. Performance Testing)} \\
frn-12 & We recommend that you provide clear instructions for use that delineate the technological features of the specific device and how to use the device. \emph{(Safety Assurance Case: D. Labeling)} \\
\bottomrule
\end{tabularx}
\end{table*}

\begin{table*}[p]
\centering\small
\caption{Label transitions from the reference run to the H100 run (T7) and to the chunk-budget 4{,}096 run (T8), over the 776 items. Rows: reference label; columns: label in the test run (``none'': the output was cut off at the output limit and has no label). Of the 34 changes on the H100, the model's own label changed in 23, the verification rule downgraded the H100 label in 8, and the reference label had been downgraded while the H100 label was not in 3; of the 6 changes at the smaller budget, 4 and 2.}
\label{tab:transitions}
\small\begin{tabular}{lrrrrcrrrr}
\toprule
& \multicolumn{4}{c}{T7 H100} & & \multicolumn{4}{c}{T8 budget 4{,}096} \\
\cmidrule(lr){2-5}\cmidrule(lr){7-10}
Reference label & Satisfied & Not sat. & Cannot det. & none & & Satisfied & Not sat. & Cannot det. & none \\
\midrule
Satisfied & 359 & 0 & 21 & 0 & & 377 & 0 & 3 & 0 \\
Not satisfied & 1 & 37 & 2 & 0 & & 0 & 39 & 1 & 0 \\
Cannot determine & 7 & 2 & 346 & 1 & & 2 & 0 & 354 & 0 \\
\bottomrule
\end{tabular}
\end{table*}

\begin{figure*}[p]
\centering
\begin{tikzpicture}
\begin{groupplot}[group style={group size=3 by 1, horizontal sep=6mm}, scale only axis, width=34mm, height=11cm, ytick={1,...,15}, y dir=reverse, ymin=0.3, ymax=15.7, yticklabel style={font=\scriptsize}, xticklabel style={font=\scriptsize}, title style={font=\small}, xmajorgrids, grid style={gray!20}, axis line style={gray!60}, ytick style={draw=none}, legend style={at={(0.5,-0.08)}, anchor=north, legend columns=3, font=\scriptsize, draw=none}]
\nextgroupplot[title={Label}, xmin=-0.01, xmax=0.25, yticklabels={Ours (self-hosted) / gemma-4-31b, Gemini API / gemini-3.8-flash, Agent Platform / gemini-3.8-flash, Gemini API / gemini-3.6-flash, Agent Platform / gemini-3.6-flash, Claude API / claude-haiku-4-5, Amazon Bedrock / claude-haiku-4-5, Claude API / claude-opus-5-5, Amazon Bedrock / claude-opus-5-5, Claude API / claude-haiku-5-5, Amazon Bedrock / claude-haiku-5-5, OpenAI API / gpt-6-astra, Azure OpenAI / gpt-6-astra, OpenAI API / gpt-6-luna, Azure OpenAI / gpt-6-luna}]
\addplot[only marks, mark=*, mark size=1.6pt, black, error bars/.cd, x dir=both, x explicit] table[x=p, y=i, x error minus=em, x error plus=ep] {
p i em ep
0.0 0.78 0.0000 0.0000
0.0 1.78 0.0000 0.0000
0.0 2.78 0.0000 0.0000
0.0 3.78 0.0000 0.0000
0.009 4.78 0.0090 0.0170
0.034 5.78 0.0250 0.0350
0.0 6.78 0.0000 0.0000
0.081 7.78 0.0440 0.0480
0.08 8.78 0.0430 0.0490
0.091 9.78 0.0420 0.0490
0.104 10.78 0.0470 0.0550
0.046 11.78 0.0320 0.0400
0.077 12.78 0.0430 0.0520
0.103 13.78 0.0460 0.0560
0.137 14.78 0.0570 0.0600
};
\addplot[only marks, mark=square*, mark size=1.6pt, blue!70!black, error bars/.cd, x dir=both, x explicit] table[x=p, y=i, x error minus=em, x error plus=ep] {
p i em ep
0.0 1.0 0.0000 0.0000
0.0 2.0 0.0000 0.0000
0.0 3.0 0.0000 0.0000
0.0 4.0 0.0000 0.0000
0.017 5.0 0.0170 0.0260
0.023 6.0 0.0230 0.0280
0.0 7.0 0.0000 0.0000
0.086 8.0 0.0400 0.0450
0.069 9.0 0.0380 0.0450
0.083 10.0 0.0430 0.0480
0.071 11.0 0.0370 0.0430
0.053 12.0 0.0320 0.0380
0.043 13.0 0.0320 0.0340
0.123 14.0 0.0520 0.0560
0.114 15.0 0.0480 0.0520
};
\addplot[only marks, mark=triangle*, mark size=1.6pt, red!75!black, error bars/.cd, x dir=both, x explicit] table[x=p, y=i, x error minus=em, x error plus=ep] {
p i em ep
0.0 1.22 0.0000 0.0000
0.014 2.22 0.0140 0.0290
0.029 3.22 0.0290 0.0420
0.0 4.22 0.0000 0.0000
0.013 5.22 0.0130 0.0210
0.026 6.22 0.0200 0.0280
0.0 7.22 0.0000 0.0000
0.077 8.22 0.0360 0.0390
0.078 9.22 0.0400 0.0450
0.094 10.22 0.0400 0.0440
0.101 11.22 0.0410 0.0500
0.059 12.22 0.0360 0.0430
0.062 13.22 0.0350 0.0370
0.107 14.22 0.0430 0.0470
0.128 15.22 0.0420 0.0490
};
\nextgroupplot[title={Cited passages}, xmin=-0.01, xmax=0.55, yticklabels={}]
\addplot[only marks, mark=*, mark size=1.6pt, black, error bars/.cd, x dir=both, x explicit] table[x=p, y=i, x error minus=em, x error plus=ep] {
p i em ep
0.0 0.78 0.0000 0.0000
0.0 1.78 0.0000 0.0000
0.0 2.78 0.0000 0.0000
0.026 3.78 0.0260 0.0340
0.023 4.78 0.0230 0.0280
0.093 5.78 0.0530 0.0560
0.006 6.78 0.0060 0.0110
0.2 7.78 0.0670 0.0700
0.241 8.78 0.0710 0.0720
0.351 9.78 0.0750 0.0750
0.367 10.78 0.0740 0.0760
0.084 11.78 0.0440 0.0470
0.109 12.78 0.0460 0.0570
0.167 13.78 0.0680 0.0730
0.193 14.78 0.0630 0.0660
};
\addplot[only marks, mark=square*, mark size=1.6pt, blue!70!black, error bars/.cd, x dir=both, x explicit] table[x=p, y=i, x error minus=em, x error plus=ep] {
p i em ep
0.0 1.0 0.0000 0.0000
0.0 2.0 0.0000 0.0000
0.0 3.0 0.0000 0.0000
0.026 4.0 0.0260 0.0340
0.031 5.0 0.0250 0.0350
0.067 6.0 0.0410 0.0490
0.009 7.0 0.0090 0.0170
0.206 8.0 0.0620 0.0650
0.161 9.0 0.0620 0.0690
0.299 10.0 0.0760 0.0800
0.31 11.0 0.0700 0.0770
0.096 12.0 0.0450 0.0480
0.08 13.0 0.0370 0.0400
0.176 14.0 0.0630 0.0710
0.203 15.0 0.0630 0.0660
};
\addplot[only marks, mark=triangle*, mark size=1.6pt, red!75!black, error bars/.cd, x dir=both, x explicit] table[x=p, y=i, x error minus=em, x error plus=ep] {
p i em ep
0.0 1.22 0.0000 0.0000
0.014 2.22 0.0140 0.0290
0.029 3.22 0.0290 0.0420
0.021 4.22 0.0210 0.0200
0.025 5.22 0.0200 0.0270
0.127 6.22 0.0600 0.0700
0.006 7.22 0.0060 0.0130
0.202 8.22 0.0610 0.0600
0.197 9.22 0.0600 0.0620
0.334 10.22 0.0720 0.0760
0.353 11.22 0.0700 0.0720
0.1 12.22 0.0460 0.0510
0.105 13.22 0.0420 0.0460
0.175 14.22 0.0610 0.0600
0.2 15.22 0.0560 0.0620
};
\nextgroupplot[title={Full output}, xmin=-0.01, xmax=1.05, yticklabels={}]
\addplot[only marks, mark=*, mark size=1.6pt, black, error bars/.cd, x dir=both, x explicit] table[x=p, y=i, x error minus=em, x error plus=ep] {
p i em ep
0.0 0.78 0.0000 0.0000
0.091 1.78 0.0420 0.0490
0.074 2.78 0.0400 0.0430
0.463 3.78 0.0740 0.0770
0.484 4.78 0.0530 0.0460
0.319 5.78 0.0930 0.0910
0.011 6.78 0.0110 0.0180
1.0 7.78 0.0000 0.0000
0.999 8.78 0.0030 0.0010
1.0 9.78 0.0000 0.0000
0.996 10.78 0.0060 0.0040
0.981 11.78 0.0100 0.0100
0.979 12.78 0.0230 0.0150
0.999 13.78 0.0030 0.0010
0.994 14.78 0.0110 0.0060
};
\addplot[only marks, mark=square*, mark size=1.6pt, blue!70!black, error bars/.cd, x dir=both, x explicit] table[x=p, y=i, x error minus=em, x error plus=ep] {
p i em ep
0.0 1.0 0.0000 0.0000
0.071 2.0 0.0370 0.0430
0.011 3.0 0.0110 0.0180
0.443 4.0 0.0790 0.0730
0.497 5.0 0.0370 0.0340
0.33 6.0 0.0800 0.0830
0.023 7.0 0.0230 0.0280
0.999 8.0 0.0030 0.0010
1.0 9.0 0.0000 0.0000
1.0 10.0 0.0000 0.0000
0.999 11.0 0.0030 0.0010
0.977 12.0 0.0180 0.0140
0.974 13.0 0.0340 0.0220
0.997 14.0 0.0040 0.0030
0.999 15.0 0.0030 0.0010
};
\addplot[only marks, mark=triangle*, mark size=1.6pt, red!75!black, error bars/.cd, x dir=both, x explicit] table[x=p, y=i, x error minus=em, x error plus=ep] {
p i em ep
0.0 1.22 0.0000 0.0000
0.183 2.22 0.0760 0.0770
0.519 3.22 0.1120 0.1140
0.407 4.22 0.0640 0.0630
0.489 5.22 0.0260 0.0280
0.361 6.22 0.0840 0.0870
0.015 7.22 0.0150 0.0190
1.0 8.22 0.0000 0.0000
0.999 9.22 0.0020 0.0010
0.998 10.22 0.0020 0.0020
0.997 11.22 0.0030 0.0030
0.989 12.22 0.0090 0.0080
0.981 13.22 0.0260 0.0160
0.998 14.22 0.0030 0.0020
0.997 15.22 0.0040 0.0030
};
\end{groupplot}
\end{tikzpicture}
\caption{Experiment 1: probability that two runs of the same prompt differ, by field and deployment, within day 1 (circles), within day 2 (squares) and across days (triangles), with 95\% percentile bootstrap intervals over the 70 prompts (Methods). Rows are in the order of Table~\ref{tab:e1}; the values repeat those of Tables~\ref{tab:e1} and~\ref{tab:e1days} and add the cited-passages field and the intervals.}
\label{fig:s1}
\end{figure*}
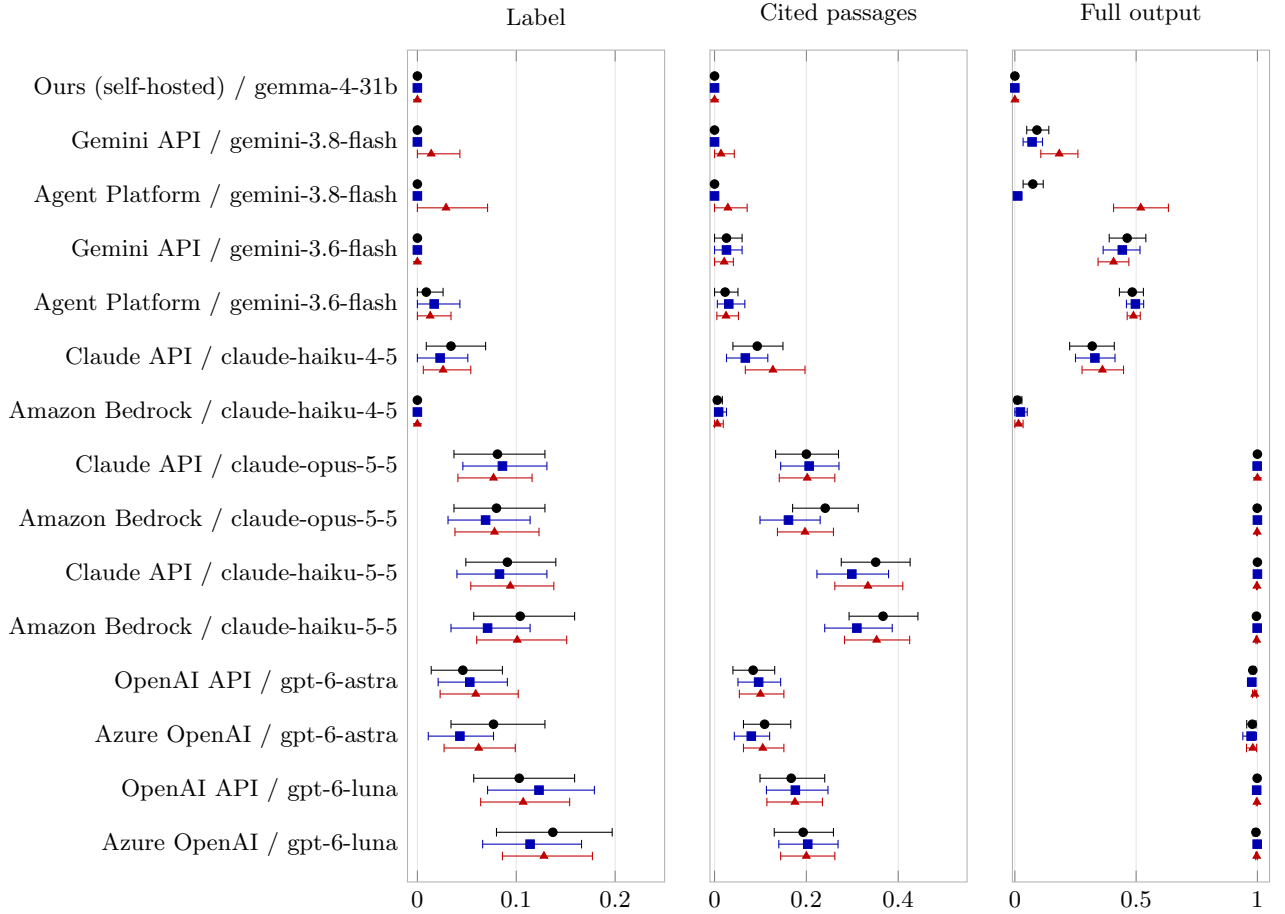

%% file: supp52.tex
\begin{sidewaystable*}
\centering\footnotesize
\caption{Experiment 1 addendum: the pipeline's own prompts sent to the commercial deployments (49 device--item pairs, the retrieved six chunks of the reference run, 7{,}200--8{,}700 characters; 5 runs on each of 14--15 October 2026, provider-default reasoning). Columns 3--8: prompts (of 49) on which the field differed in at least one of the five runs of day 1; then the probability that two runs differ in label (day 1, with 95\% bootstrap interval) and in label and output within day 1 / within day 2 / across days. The pipeline's own judgment stage returned byte-identical output for these prompts across nine runs (Table~\ref{tab:e3}). Two calls of \texttt{claude-opus-5-5} through the Claude API on day 2 failed and are omitted.}
\label{tab:e3prompts}
\begin{tabular}{llrrrrrrlll}
\toprule
Route & Model & Label & Files & Passages & Quotes & Comment & Output & $P$(label), day 1 & $P$(label) d1 / d2 / across & $P$(output) d1 / d2 / across \\
\midrule
Gemini API & \texttt{gemini-3.8-flash} & 0 & 0 & 0 & 0 & 0 & 0 & 0.000 --- & 0.000 / 0.000 / 0.000 & 0.000 / 0.000 / 0.000 \\
Agent Platform & \texttt{gemini-3.8-flash} & 0 & 0 & 0 & 0 & 1 & 1 & 0.000 --- & 0.000 / 0.000 / 0.000 & 0.012 / 0.008 / 0.155 \\
Gemini API & \texttt{gemini-3.6-flash} & 0 & 0 & 0 & 0 & 0 & 0 & 0.000 --- & 0.000 / 0.000 / 0.000 & 0.000 / 0.000 / 0.000 \\
Agent Platform & \texttt{gemini-3.6-flash} & 1 & 3 & 6 & 7 & 48 & 48 & 0.012 [0.000, 0.037] & 0.012 / 0.012 / 0.010 & 0.535 / 0.498 / 0.513 \\
Claude API & \texttt{claude-haiku-4-5} & 0 & 2 & 3 & 4 & 5 & 5 & 0.000 --- & 0.000 / 0.000 / 0.000 & 0.063 / 0.031 / 0.046 \\
Amazon Bedrock & \texttt{claude-haiku-4-5} & 0 & 0 & 0 & 0 & 0 & 0 & 0.000 --- & 0.000 / 0.016 / 0.008 & 0.000 / 0.243 / 0.139 \\
Claude API & \texttt{claude-opus-5-5} & 4 & 13 & 22 & 30 & 49 & 49 & 0.033 [0.008, 0.065] & 0.033 / 0.025 / 0.027 & 0.998 / 1.000 / 1.000 \\
Amazon Bedrock & \texttt{claude-opus-5-5} & 2 & 15 & 23 & 27 & 49 & 49 & 0.020 [0.000, 0.053] & 0.020 / 0.041 / 0.035 & 1.000 / 1.000 / 0.998 \\
Claude API & \texttt{claude-haiku-5-5} & 2 & 29 & 41 & 44 & 49 & 49 & 0.016 [0.000, 0.041] & 0.016 / 0.041 / 0.029 & 1.000 / 1.000 / 1.000 \\
Amazon Bedrock & \texttt{claude-haiku-5-5} & 4 & 29 & 38 & 42 & 49 & 49 & 0.041 [0.008, 0.086] & 0.041 / 0.033 / 0.042 & 1.000 / 1.000 / 1.000 \\
OpenAI API & \texttt{gpt-6-astra} & 3 & 8 & 14 & 22 & 47 & 48 & 0.029 [0.000, 0.065] & 0.029 / 0.033 / 0.037 & 0.926 / 0.953 / 0.948 \\
Azure OpenAI & \texttt{gpt-6-astra} & 3 & 8 & 19 & 28 & 45 & 48 & 0.029 [0.000, 0.065] & 0.029 / 0.025 / 0.039 & 0.922 / 0.908 / 0.945 \\
OpenAI API & \texttt{gpt-6-luna} & 7 & 25 & 30 & 34 & 49 & 49 & 0.082 [0.033, 0.143] & 0.082 / 0.110 / 0.102 & 0.978 / 0.992 / 0.990 \\
Azure OpenAI & \texttt{gpt-6-luna} & 7 & 29 & 34 & 39 & 49 & 49 & 0.069 [0.025, 0.122] & 0.069 / 0.098 / 0.083 & 0.990 / 0.984 / 0.985 \\
\bottomrule
\end{tabular}
\end{sidewaystable*}

\begin{sidewaystable*}
\centering\footnotesize
\caption{Experiment 1 addendum: reasoning minimised (the 70 prompts of Table~\ref{tab:e1}, 5 runs on each of 14--15 October 2026). The eight deployments whose interface allows reasoning to be reduced below the default were re-measured with the lowest documented setting; all other settings as in Table~\ref{tab:e1routes}. Columns as in Supplementary Table~\ref{tab:e3prompts}; ``Default'' repeats the day-1 label and output counts at provider-default reasoning (Table~\ref{tab:e1}). Reasoning tokens were reported on none of the 5{,}600 calls except 56 calls of \texttt{gemini-3.8-flash} through Agent Platform. \texttt{gpt-6-luna} with reasoning effort none reached the 16{,}384-token output limit without a parsable answer on 24 of 700 calls through the OpenAI API and 29 through Azure (each such call counts as differing in every field); \texttt{claude-opus-5-5} and \texttt{gpt-6-astra} reject the setting and are not listed.}
\label{tab:e4reason}
\begin{adjustbox}{max width=\linewidth}
\begin{tabular}{lllrrrrrrllll}
\toprule
Route & Model & Setting & Label & Files & Passages & Quotes & Comment & Output & Default (label, output) & $P$(label), day 1 & $P$(label) d1 / d2 / across & $P$(output) d1 / d2 / across \\
\midrule
Gemini API & \texttt{gemini-3.8-flash} & thinking level LOW & 4 & 3 & 6 & 16 & 68 & 68 & 0, 14 & 0.023 [0.006, 0.046] & 0.023 / 0.037 / 0.032 & 0.451 / 0.451 / 0.479 \\
Agent Platform & \texttt{gemini-3.8-flash} & thinking level LOW & 0 & 0 & 0 & 0 & 0 & 0 & 0, 11 & 0.000 --- & 0.000 / 0.000 / 0.014 & 0.000 / 0.006 / 0.143 \\
Gemini API & \texttt{gemini-3.6-flash} & thinking level MINIMAL & 7 & 9 & 15 & 20 & 54 & 54 & 0, 52 & 0.049 [0.017, 0.086] & 0.049 / 0.037 / 0.039 & 0.399 / 0.400 / 0.358 \\
Agent Platform & \texttt{gemini-3.6-flash} & thinking level MINIMAL & 5 & 6 & 13 & 18 & 64 & 64 & 1, 61 & 0.040 [0.009, 0.077] & 0.040 / 0.040 / 0.033 & 0.500 / 0.486 / 0.532 \\
Claude API & \texttt{claude-haiku-5-5} & thinking disabled & 24 & 36 & 46 & 53 & 70 & 70 & 13, 70 & 0.164 [0.109, 0.224] & 0.164 / 0.149 / 0.154 & 0.999 / 0.997 / 0.994 \\
Amazon Bedrock & \texttt{claude-haiku-5-5} & thinking disabled & 20 & 26 & 42 & 52 & 70 & 70 & 14, 70 & 0.134 [0.086, 0.189] & 0.134 / 0.159 / 0.153 & 1.000 / 0.996 / 0.998 \\
OpenAI API & \texttt{gpt-6-luna} & reasoning effort none & 42 & 25 & 42 & 47 & 70 & 70 & 14, 70 & 0.299 [0.240, 0.361] & 0.299 / 0.323 / 0.350 & 1.000 / 1.000 / 0.999 \\
Azure OpenAI & \texttt{gpt-6-luna} & reasoning effort none & 43 & 28 & 42 & 45 & 70 & 70 & 18, 70 & 0.344 [0.273, 0.410] & 0.344 / 0.307 / 0.314 & 1.000 / 1.000 / 1.000 \\
\bottomrule
\end{tabular}
\end{adjustbox}
\end{sidewaystable*}